\documentclass{article}

\usepackage{fancyhdr}
\usepackage{iftex}

\ifPDFTeX
  \usepackage[T1]{fontenc}
  \usepackage[utf8]{inputenc}
\else
\fi

\usepackage{report}

\usepackage{array}
\usepackage{booktabs}

\usepackage{tabularx}
\usepackage{xcolor}
\usepackage{colortbl}
\usepackage[most]{tcolorbox}

\newtcolorbox{casebox}[2][]{
  enhanced,
  breakable,
  colback=gray!2,
  colframe=black!18,
  boxrule=0.4pt,
  arc=2pt,
  left=4pt,right=4pt,top=2pt,bottom=2pt,
  fonttitle=\bfseries\small,
  title={#2},
  #1
}

\newcolumntype{Y}{>{\raggedright\arraybackslash}X}
\renewcommand{\arraystretch}{1.18}
\newcolumntype{C}{>{\centering\arraybackslash}X} 

\usepackage{graphicx}
\usepackage{subfigure}
\usepackage{booktabs} 

\usepackage{pifont}
\usepackage{soul}
\RequirePackage{natbib}
\usepackage{makecell}
\usepackage{graphicx} 

\usepackage{url}            
\usepackage{booktabs}       
\usepackage{amsfonts}       
\usepackage{fancyhdr}       

\usepackage{nicefrac}       
\usepackage{graphicx}
\usepackage{pifont}
\usepackage{multirow}
\definecolor{darkmagenta}{rgb}{0.56, 0.0, 1.0}
\definecolor{softyellow}{rgb}{1.0, 0.92, 0.3} 
\definecolor{LightAquamarine}{rgb}{0.75, 1.0, 0.8} 
\definecolor{FireBrick}{RGB}{178,34,34}
\definecolor{MediumPurple}{RGB}{147,112,219}

\definecolor{uclablue}{rgb}{0.15, 0.45, 0.68}
\hypersetup{
    breaklinks,
    colorlinks=true,
    citecolor={darkmagenta},
    linkcolor={uclablue},
    urlcolor={uclablue}
}
\usepackage{wrapfig}
\usepackage{float}
\usepackage{subcaption}
\usepackage{placeins}
\usepackage{lipsum} 
\usepackage{tcolorbox}
\usepackage{amsmath}
\usepackage{amssymb}
\usepackage{tcolorbox}
\usepackage{fancyvrb}
\usepackage{verbatim}

\usepackage{utfsym}
\usepackage{fontawesome}
\usepackage{xspace}
\usepackage{enumitem}
\usepackage{multirow} 
\tcbuselibrary{breakable}
\usepackage{enumitem}
\usepackage{colortbl}
\usepackage{fancyhdr}
\usepackage{tabularx}
\usepackage{tcolorbox}
\usepackage{transparent}

\definecolor{njuPurple}{RGB}{220,205,230}     
\definecolor{njuPurpleLight}{RGB}{250,245,252}   

\newtcolorbox{abstractbox}{
    colback=njuPurpleLight,   
    colframe=njuPurple,       
    boxrule=1pt,              
    arc=4mm,                  
    left=8pt,                 
    right=8pt,                
    top=8pt,                  
    bottom=8pt,               
    opacityback=0.95
}

\usepackage{listings}
\usepackage[table]{xcolor}
\definecolor{oursblue}{HTML}{EAF3FF}
\definecolor{gainMint}{HTML}{DDF6E6}
\definecolor{dropPink}{HTML}{F8DCDC}

\newcommand{\drop}[1]{%
  \textsuperscript{\scriptsize\setlength{\fboxsep}{0.4pt}\colorbox{dropPink}{\textcolor{black}{$\downarrow$#1}}}%
}

\newcommand{\gain}[1]{%
  \textsuperscript{\scriptsize\setlength{\fboxsep}{0.4pt}\colorbox{gainMint}{\textcolor{black}{$\uparrow$#1}}}%
}

\newcommand{\tagbox}[2]{%
  \tcbox[
    on line,
    box align=base,
    boxsep=0pt,
    left=2.6pt,
    right=2.6pt,
    top=0.5pt,
    bottom=0.5pt,
    arc=2pt,
    outer arc=2pt,
    boxrule=0.35pt,
    colback=#1,
    colframe=#1!65!black,
    coltext=black,
    fontupper=\scriptsize\bfseries
  ]{#2}%
}

\definecolor{tagStep}{RGB}{226,245,238}
\definecolor{tagPRM}{RGB}{232,235,255}
\definecolor{tagTrace}{RGB}{226,245,226}
\definecolor{tagType}{RGB}{255,226,226}
\definecolor{tagModal}{RGB}{225,245,255}
\definecolor{tagError}{RGB}{255,226,226}
\definecolor{tagRoot}{RGB}{255,238,210}
\definecolor{tagDep}{RGB}{240,235,255}
\definecolor{tagOurs}{RGB}{242,240,255}
\definecolor{tagSection}{RGB}{206,245,238}

\tcbuselibrary{listings,breakable,skins}

\lstdefinestyle{promptstyle}{
  basicstyle=\ttfamily\footnotesize,
  breaklines=true,
  breakatwhitespace=false,
  columns=fullflexible,
  keepspaces=true,
  showstringspaces=false,
  tabsize=2
}

\newtcblisting{promptbox}[2][]{
  title={#2},
  colback=white,
  colframe=black,
  boxrule=0.8pt,
  arc=1pt,
  outer arc=1pt,
  left=1.0em,
  right=1.0em,
  top=0.8em,
  bottom=0.8em,
  fonttitle=\bfseries,
  listing only,
  listing style=promptstyle,
  breakable,
  enhanced,
  #1
}

\lstdefinestyle{casestyle}{
  basicstyle=\ttfamily\footnotesize,
  breaklines=true,
  breakatwhitespace=false,
  columns=fullflexible,
  keepspaces=true,
  showstringspaces=false,
  tabsize=2
}

\newtcblisting{casetbox}[2][]{
  title={#2},
  colback=white,
  colframe=black,
  boxrule=0.8pt,
  arc=1pt,
  outer arc=1pt,
  left=1.0em,
  right=1.0em,
  top=0.8em,
  bottom=0.8em,
  fonttitle=\bfseries,
  listing only,
  listing style=promptstyle,
  breakable,
  enhanced,
  #1
}

\usepackage{xcolor}
\usepackage{tcolorbox}
\usepackage{tabularx}
\usepackage{enumitem}
\usepackage{makecell}

\tcbuselibrary{skins, breakable}

\definecolor{caseframe}{RGB}{55, 55, 55}
\definecolor{casebg}{RGB}{245, 245, 245}

\definecolor{errbg}{RGB}{255, 248, 252}
\definecolor{errframe}{RGB}{245, 180, 210}

\definecolor{normbg}{RGB}{235, 250, 242}
\definecolor{normframe}{RGB}{95, 180, 140}
\newtcolorbox{trajcase}[1]{
  enhanced, breakable,
  colframe=caseframe, colback=casebg,
  boxrule=0.6pt, arc=2pt,
  fonttitle=\bfseries\small,
  title={#1},
  top=4pt, bottom=4pt, left=4pt, right=4pt,
}

\newtcolorbox{normalspan}[1]{
  enhanced, breakable,
  colframe=normframe, colback=normbg,
  boxrule=0.4pt, arc=1.5pt,
  fonttitle=\bfseries\footnotesize,
  title={#1},
  top=3pt, bottom=3pt, left=3pt, right=3pt,
  before skip=4pt, after skip=4pt,
}

\newtcolorbox{errorspan}[2]{
  enhanced, breakable,
  colframe=errframe, colback=errbg,
  boxrule=0.4pt, arc=1.5pt,
  fonttitle=\bfseries\footnotesize,
  title={#1 \normalfont\itshape\footnotesize(#2)},
  top=3pt, bottom=3pt, left=3pt, right=3pt,
  before skip=4pt, after skip=4pt,
}

\title{Where Do Deep-Research Agents Go Wrong? Span-Level Error Localization in Agent Trajectories}

\author{
Jiaming Wang$^{1*}$,
Ziteng Feng$^{1*}$,
Jiangtao Wu$^{1}$,
Ruihao Li$^{1}$,
Qianqian Xie$^{1}$,\\
\vspace{1mm}
Yuxiang Ren$^{1}$, 
He Zhu$^{1}$,
Xueming Han$^{2}$,
Fanyu Meng$^{2}$,
Junlan Feng$^{2}$,
Jiaheng Liu$^{1,\dagger}$\\
\vspace{4mm}
{\small $^1$ \textbf{NJU-LINK Team, Nanjing University}} \quad
{\small $^2$  \textbf{JIUTIAN Research}}  
\\
\vspace{2mm}
\texttt{jiaming\_wang@smail.nju.edu.cn}
\quad\quad\quad
\texttt{liujiaheng@nju.edu.cn} \\
}

\begin{document}
\maketitle

\renewcommand{\thefootnote}{\fnsymbol{footnote}}
\footnotetext[1]{Equal Contribution.}
\footnotetext[2]{Corresponding Author.}
\renewcommand{\thefootnote}{\arabic{footnote}}

\begin{abstractbox}
\begin{center}
\textbf{\Large Abstract}
\end{center}
\noindent
    Deep-research agents solve tasks through long trajectories of search, tool use, evidence inspection, and answer synthesis. Evaluation based on final answers shows whether an agent succeeds, but not which parts of the trajectory make the answer unreliable. We study span-level error localization for deep-research agents. We collect 2,790 real trajectories from two agent frameworks, three backbone models, and three benchmarks, convert raw logs into semantic spans, and annotate harmful error spans through LLM-assisted expert review. From these annotations, we build \textsc{TELBench}\footnote{\url{https://huggingface.co/datasets/NJU-LINK/TELBench}}, a 1,000-instance benchmark for identifying error spans among normal exploration, failed searches, tentative hypotheses, and harmless noise. We further propose \textsc{DRIFT}~\footnote{\url{https://github.com/NJU-LINK/DRIFT}} , a claim-centric auditing framework that tracks agent claims, checks their support in trajectory evidence, and marks spans where unsupported or conflicting claims affect the answer path. Experiments across model families and auditing frameworks show that \textsc{DRIFT} improves span-level error localization and first-error accuracy by up to 30 percentage points. Our work provides a process-level view of reliability in deep-research agents.
    
\end{abstractbox}

\section{Introduction}

A deep-research trajectory is better viewed as a recorded decision process than as a single input-output computation~\citep{deshpande2025trailtracereasoningagentic, yao2023reactsynergizingreasoningacting}. It gradually forms claims about entities, constraints, sources, intermediate candidates, and final conclusions, and later spans often reuse earlier claims as if they were established facts. Its log records not only external actions, but also the evolution of commitments: which claims are introduced, what evidence supports them, and where they are later reused~\citep{qin2023toolllm, Kim2025BeyondTF}.

The difficulty is that the harmful step is often not the visibly wrong final answer, but an earlier commitment that later spans inherit without revalidation~\citep{ICLR2024_aca97732,zhang2025agentcausestaskfailures}. Evaluation based on final answers can tell us whether an agent succeeded, but not which part of the trajectory made the result unreliable~\citep{chen2026seeingelephantbenchmarkfailure, NEURIPS2023_cd86a305}. Raw logs contain the needed evidence, but they are long, heterogeneous, and framework-specific. Directly asking an LLM to find errors in the full trajectory is also unstable: it may mistake benign exploration for an error, over-focus on the final answer, or miss an early unsupported commitment that later shapes the solution. The key diagnostic question is therefore not only which span appears wrong, but which unsupported claim first became consequential and which later spans rely on it.

To study this problem, we represent agent trajectories as ordered semantic spans. Semantic spans provide an analysis unit that is coarser than raw events but still precise enough to localize the first harmful commitment. We collect 2,790 real agent trajectories from two agent frameworks, three backbone models, and three challenging deep-research benchmarks, convert them into semantic spans, and annotate harmful errors through dual human annotation and review~\citep{dligach-palmer-2011-reducing, artstein-poesio-2008-survey}. Based on these annotations, we construct \textsc{TELBench}, a benchmark for span-level trajectory error localization~\citep{tyen-etal-2024-llms}. Given only the question and ordered raw span texts, a model must identify error and non-error spans such as benign exploration or noise.

We further propose \textsc{DRIFT}, a claim-centric multi-agent auditing framework for trajectory error localization. Rather than scoring spans independently, \textsc{DRIFT} audits the claims that an agent forms and uses throughout the trajectory~\citep{thorne-etal-2018-fever}. A Claim Keeper reads the full trajectory and maintains a claim ledger, recording when each claim is introduced, when it becomes consequential, and which later spans depend on it. A Support Seeker checks whether key claims are directly supported, weakly supported, missing support, or contradicted by trajectory evidence. Specialist Auditors then perform skill-routed checks for entity, constraint, evidence, retrieval, compute, and process claims~\citep{NEURIPS2023_cd86a305}. Finally, a Dependency Tracer backtraces unsupported or conflicting claims to distinguish errors and non-errors.

Our contributions are threefold:
\begin{itemize}
    \item \textbf{A large-scale trajectory corpus.}
    We collect and annotate 2,790 real deep-research agent trajectories across multiple frameworks, models, and benchmarks, providing span-level analysis.

    \item \textbf{A process-level localization benchmark.}
    We introduce \textsc{TELBench}, a benchmark that evaluates whether models can localize harmful error spans from ordered trajectory evidence.

    \item \textbf{A claim-centric auditing framework.}
    We propose \textsc{DRIFT}, an auditing agent that reasons over claim support and dependency structure, outperforming direct full-context LLM prompting on trajectory error localization.
\end{itemize}
\begin{table*}[htbp]
\centering
\small
\caption{
Comparison with process-level reasoning and agent trace error localization benchmarks. For \textsc{TELBench}, items are reported as trajectories / spans. Avg. Len. denotes the average number of reasoning steps, trace steps, or spans per item when such statistics are publicly available. \textsc{TELBench} targets deep-research agent trajectories with span-level error labels, earliest harmful span localization, and dependency-aware error propagation.
}
\setlength{\tabcolsep}{5.0pt}
\renewcommand{\arraystretch}{1.08}

\resizebox{\textwidth}{!}{
\begin{tabular}{@{}lllclc@{}}
\toprule
\textbf{Benchmark} &
\textbf{Task} &
\textbf{Items} &
\textbf{Trace Type} &
\textbf{Avg. Len.} &
\textbf{Eval. Dimensions} \\
\midrule

ProcessBench &
Process Error ID &
3,400 &
Math CoT &
7.56 steps &
\tagbox{tagStep}{Step Err.} \tagbox{tagPRM}{PRM} \\

PRMBench &
PRM Diagnosis &
6,216 &
Reasoning Path &
13.43 steps &
\tagbox{tagPRM}{PRM} \tagbox{tagType}{Type} \\

DeltaBench &
Long-CoT Error ID &
1,236 &
Long CoT &
-- &
\tagbox{tagSection}{Section Err.} \\

VisualProcessBench &
Multimodal PRM &
2,866 &
VLM CoT &
9.40 steps &
\tagbox{tagPRM}{PRM} \tagbox{tagModal}{Multi-modal} \\

AgentProcessBench &
Agent Process Eval. &
1,000 &
Tool Agent Trace &
8.51 steps &
\tagbox{tagStep}{Step Err.} \tagbox{tagTrace}{Trace Diag.} \\

TRAIL &
Agent Issue Loc. &
148 &
Agent Trace &
-- &
\tagbox{tagTrace}{Trace Diag.} \\

\midrule

\rowcolor{tagOurs}
\textbf{\textsc{TELBench} (Ours)} &
\textbf{DR Error Loc.} &
\textbf{1,000} &
\textbf{DR Agent Trace} &
\textbf{11.95 spans} &
\tagbox{tagError}{Span Err.}
\tagbox{tagRoot}{First Err.} \\

\bottomrule
\end{tabular}
}
\label{tab:related-error-localization}
\end{table*}

\section{Related Work}

\paragraph{Deep-research systems and outcome-level evaluation.}
Recent agent benchmarks, such as GAIA, BrowseComp, WebArena, and OSWorld, have shifted focus from static QA to long-horizon tasks including web navigation and tool use \citep{mialon2023gaiabenchmarkgeneralai, wei2025browsecomp, zhou2024webarenarealisticwebenvironment, xie2024osworldbenchmarkingmultimodalagents}. Furthermore, newer benchmarks like DeepResearch Bench, DeepResearchGym, LiveResearchBench, and DRBench emphasize rubric-based assessment of citation-grounded reports \citep{du2025deepresearchbenchcomprehensivebenchmark, coelho2025deepresearchgymfreetransparentreproducible, wang2026liveresearchbenchlivebenchmarkusercentric, abaskohi2026drbenchrealisticbenchmarkenterprise}. Despite improving realism, these evaluations remain primarily outcome-centered: they assess final task completion but fail to localize where a research trajectory first becomes unreliable. In contrast, TELBench evaluates the process itself by segmenting trajectories into semantic spans, requiring models to pinpoint harmful errors within ordered evidence.

\paragraph{Process-level evaluation and trajectory diagnosis.}
Moving beyond outcome-level metrics, recent works evaluate intermediate reasoning and agent traces. Frameworks such as ProcessBench, PRMBench, Delta-Bench, VisualProcessBench, AgentProcessBench, and TRACE focus on step-level or tool-use errors~\citep{zheng-etal-2025-processbench, song2025prmbenchfinegrainedchallengingbenchmark, he2025largelanguagemodelsdetect, wang2025visualprm, fan2026agentprocessbenchdiagnosingsteplevelprocess, Kim2025BeyondTF}, while MAST, TRAIL, AgentRx, and CodeTracer address failure localization and trace debugging~\citep{cemri2025multiagentllmsystemsfail, deshpande2025trailtracereasoningagentic, barke2026agentrxdiagnosingaiagent, li2026codetracertraceableagentstates}. While valuable (Table~\ref{tab:related-error-localization}), prior signals are mostly built for shorter or more structured settings such as math reasoning, VLM reasoning~\citep{wei2026agentic}, coding traces, and API workflows. Deep-research trajectories are longer and noisier, mixing useful exploration, weak evidence, failed searches, and harmful mistakes. \textsc{TELBench} focuses on this setting by evaluating whether models can localize harmful error spans from ordered semantic spans, rather than only judging final answers or overall trajectory quality.

\section{Dataset}
\subsection{Full Dataset Pipeline}
\begin{figure*}[tb!]
  \centering
  \includegraphics[width=1\textwidth]{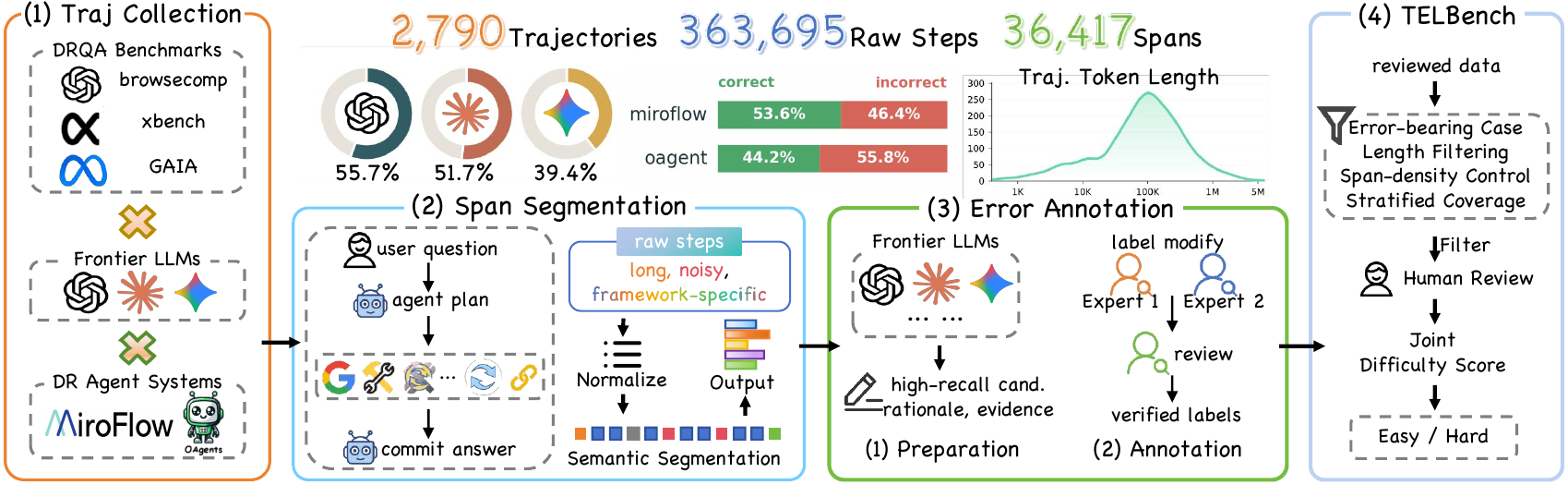}
  \caption{Data curation pipeline for \textsc{TELBench}, covering trajectory collection, log normalization, semantic-span segmentation, LLM-assisted candidate labeling, and expert-verified error annotation.}
  \label{fig:data_pipeline}
\end{figure*}
\paragraph{Trajectory collection.}
Figure~\ref{fig:data_pipeline} summarizes the full data curation pipeline from trajectory collection to span segmentation and expert-verified error annotation. We collect trajectories from three public deep-research benchmarks: GAIA-val~\cite{mialon2023gaiabenchmarkgeneralai}, XBench~\cite{chen2025xbench}, and BrowseComp-test~\cite{wei2025browsecomp}. To avoid BrowseComp dominating the corpus, we downsample it to 200 tasks, resulting in 465 tasks. For each task, we run three frontier base models, GPT-5~\citep{openai2025gpt5}, Gemini-2.5-Pro~\citep{googledeepmind2025gemini25pro}, and Claude-Sonnet-4.5~\citep{anthropic2025claudesonnet45}, under two representative agent frameworks, MiroFlow~\cite{2025mirothinker} and OAgent~\cite{zhu2025oagentsempiricalstudybuilding}. This produces 2,790 long-form agent trajectories.

\paragraph{Span segmentation.}
Raw trajectories are too long and framework-specific for direct trajectory comparison. They contain low-level artifacts such as tool retries, usage records, message wrappers, and framework-specific scheduling. We therefore convert each trajectory into semantic spans, where each span corresponds to a contiguous segment of execution around a locally coherent objective, such as planning, retrieval, verification, comparison, or finalization. We first normalize framework-specific logs into unified execution-unit sequences. For event-ordered logs, we preserve the original order after folding tool calls with their results; for nested multi-agent traces, we reconstruct semantic execution order by expanding subagent actions and treating manager-level messages mainly as contextual summaries. We then segment the unit sequence using changes in search target, candidate set, time scope, verification criterion, or reasoning objective as boundary signals. Query rewrites, retries, and adjacent evidence collection under the same local goal are kept within the same span. Automatically flagged abnormal cases and stratified samples across framework, model, benchmark, outcome, and length are further audited with LLM assistance, with final boundary overrides made only after human inspection.

\paragraph{Error span annotation.}
We annotate each trajectory at the semantic-span level. Each span receives a binary label, \emph{error} or \emph{non-error}. An error span introduces, relies on, amplifies, or finalizes a mistaken, unsupported, contradicted, or prematurely committed judgment that affects the answer path. Normal exploration, failed searches, tentative hypotheses, recovered mistakes, and tool noise are not labeled as errors by themselves. To improve coverage and reliability, we use an LLM-assisted expert annotation pipeline. For each trajectory, two independent LLM annotators from different frontier model families first propose high-recall candidate error spans with rationales and evidence references. These proposals are then validated by two expert annotators sampled from a pool of seven annotators experienced with agent systems, browsing behavior, and tool-use failures. Experts inspect the full trajectory, verify each proposed error against trajectory evidence, revise or add labels when necessary, and adjudicate disagreement, low-confidence, and boundary cases. Overall, seven expert annotators each spent over 300 hours on trajectory reading, evidence checking, label revision, and adjudication.
\paragraph{Mechanism labels.}
After binary error span labels are finalized, we add mechanism labels for analysis. Every span receives one operation-stage label, describing what the agent is doing in the process, using an eight-stage schema: planning, retrieval, source verification, extraction, computation, decision-making, recovery, and finalization. Every error span additionally receives exactly one primary-fault label, describing why the span is erroneous; non-error spans receive no fault label. The error-fault taxonomy is induced from the annotated data: three frontier LLMs generate free-form rationales for error spans, cleaned rationale keys are clustered through a hierarchical map-reduce induction process, and the resulting candidates are manually normalized into 18 primary faults grouped into six fault families. The final taxonomy is then mapped back to all error spans. Detailed construction procedures are provided in Appendix~\ref{app:taxonomy}. These mechanism labels are used only for analysis; evaluation inputs contain only the question and ordered span text, not stage labels, fault labels, judge results, or gold annotations.

\subsection{\textsc{TELBench}}
\paragraph{Design goal.}
The goal of \textsc{TELBench} is not to collect trajectories with incorrect final answers, but to build a diagnostic test set for span-level error localization. We therefore require each instance to satisfy three criteria: the error must be verifiable from trajectory-internal evidence, the span boundary must be stable enough for evaluation, and the trajectory must contain sufficient benign behavior as distractors, such as normal search, tentative hypotheses, failed exploration, or tool noise.

\paragraph{Candidate filtering.}
Starting from the annotated 2,790-trajectory corpus, we identify 1,890 trajectories with at least one span-level error, accounting for 67.7\% of the corpus. These trajectories form the initial candidate pool. We do not directly use all error-bearing trajectories because real agent logs may contain missing records, incomplete tool outputs, degenerate short runs, unverifiable error sources, or overrepresented error patterns. We therefore filter and review the pool to retain instances with clear error boundaries, trajectory-internal evidence, stable semantic-span segmentation, and enough non-error spans to make localization non-trivial. This yields 1,000 verified instances, each labeled at the semantic-span level as \emph{error} or \emph{non-error}.

\paragraph{Difficulty split.}
To evaluate both direct and subtle localization cases, we divide Verified-1K into 600 easy and 400 hard instances. Easy instances typically contain more direct error evidence, shorter trajectories, or fewer distracting spans. Hard instances involve longer trajectories, sparser or more implicit errors, more benign exploration as distractors, and challenging patterns such as evidence overclaim, constraint miss, and candidate confusion. The final test set contains an average of 11.95 semantic spans per trajectory.

\subsection{Mechanism Analysis}
\begin{figure*}[tb!]
  \centering
  \includegraphics[width=1\textwidth]{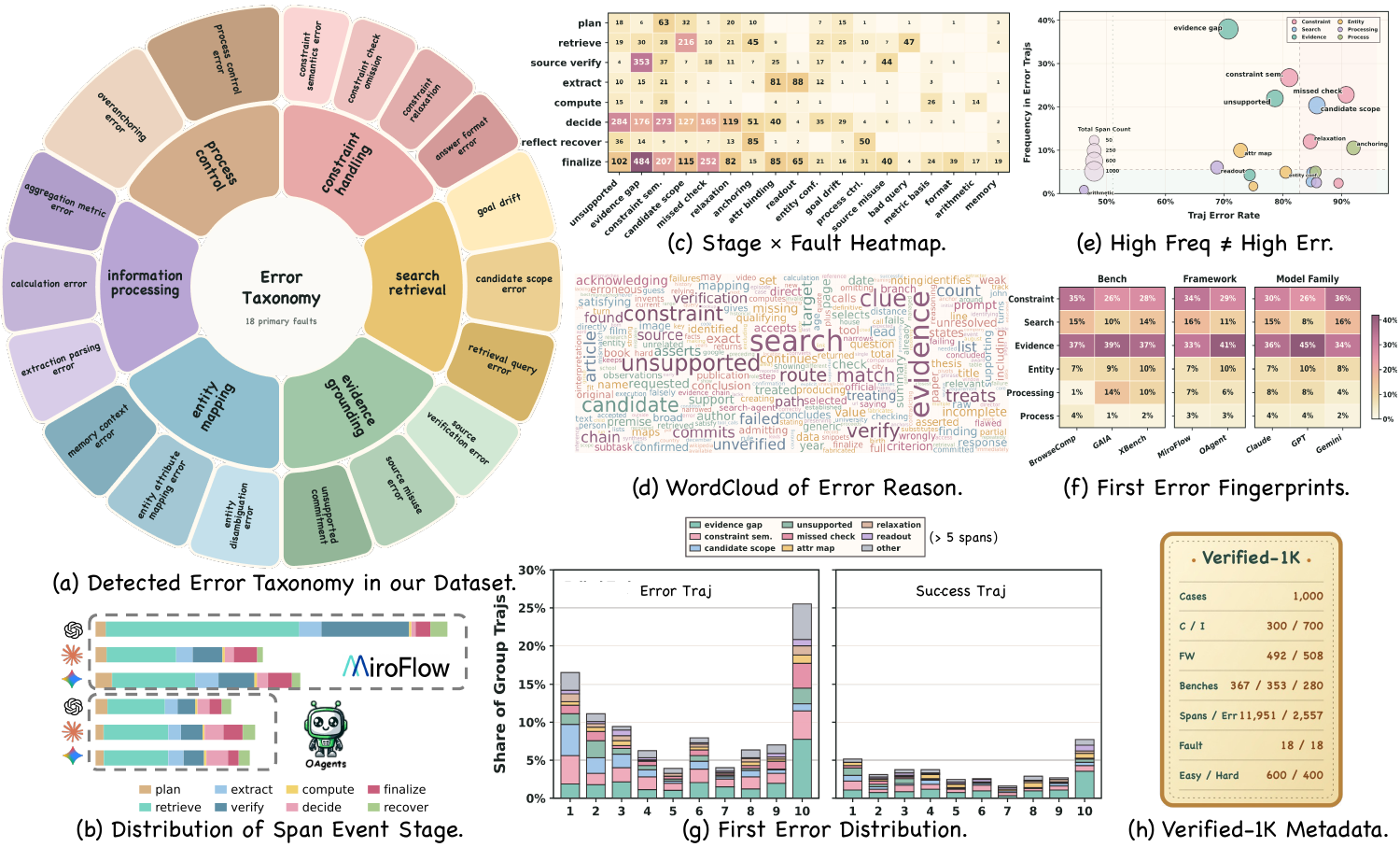}
  \caption{Mechanism analysis of annotated \textsc{TELBench} trajectories, showing error families, workflow-stage distributions, first-error patterns across settings, temporal positions, and Verified-1K coverage.}
  \label{fig:mechanism_analysis}
\end{figure*}

\paragraph{Process errors are not reducible to final outcomes.}
Before analyzing specific mechanisms, we first check whether process errors are equivalent to final-answer failure. As shown in Appendix~\ref{app:basic-analysis}, most failed trajectories contain at least one annotated error span, while 36.9\% of successful trajectories also contain process errors. This shows that span-level errors are closely related to final failure, but are not equivalent to final-answer correctness. Using the verified error-span labels and mechanism labels above, we therefore analyze trajectory failures along two axes: where an error occurs in the agent workflow and what mechanism causes it.

\paragraph{Error mechanisms are stage-structured.}
Figure~\ref{fig:mechanism_analysis}(a) shows the induced fault taxonomy, with 18 primary faults grouped into six families. Figure~\ref{fig:mechanism_analysis}(b) shows the operation-stage composition of trajectories across frameworks and model families. Figure~\ref{fig:mechanism_analysis}(c) aligns error spans with workflow stages, revealing that fault types are highly stage-dependent: candidate-scope errors concentrate in retrieval, evidence failures cluster around verification and finalization, and constraint-related errors appear more often around decision-making. The word cloud in Figure~\ref{fig:mechanism_analysis}(d) further summarizes the free-text rationales, showing recurring mechanisms such as evidence gaps, unsupported claims, search failures, and constraint misuse. Raw error counts are also affected by how often each stage appears. Appendix~\ref{app:stage-risk} normalizes errors by stage frequency and shows that retrieval is frequent but relatively low-risk, whereas decision-making and finalization have much higher normalized error rates.

\paragraph{Failure mechanisms vary across settings.}
Figure~\ref{fig:mechanism_analysis}(e) shows that fault frequency alone does not explain trajectory failure. Although evidence gaps are the most frequent error type, trajectories containing them are less likely to fail than trajectories containing several rarer faults. In contrast, missed checks, candidate-scope errors, anchoring, and constraint-semantic errors appear less often but are associated with a higher probability of trajectory failure. Figure~\ref{fig:mechanism_analysis}(f) further shows that first-error mechanisms are not uniform across settings. Across benchmarks, evidence and constraint errors dominate, but GAIA shifts noticeably toward processing errors, suggesting more failures after information has already been collected. Across frameworks, OAgent has a stronger evidence-error fingerprint than MiroFlow, while MiroFlow shows relatively more constraint and search-related first errors. Across model families, GPT is most evidence-heavy, Gemini is most constraint-heavy, and Claude is more balanced across the two. Figure~\ref{fig:mechanism_analysis}(g) adds the temporal view: failed trajectories place substantially more first errors in the earliest and latest position bins, while successful trajectories contain far fewer committed error starts. Figure~\ref{fig:mechanism_analysis}(h) summarizes the Verified-1K subset used for \textsc{TELBench}, while the other analyses are conducted on the full 2,790-trajectory annotated corpus. Qualitative examples are provided in Appendix~\ref{sec:case-study}.

\section{DRIFT: Claim-Centric Trajectory Auditing}

\begin{figure*}[!t]
  \centering
  \includegraphics[width=1\textwidth]{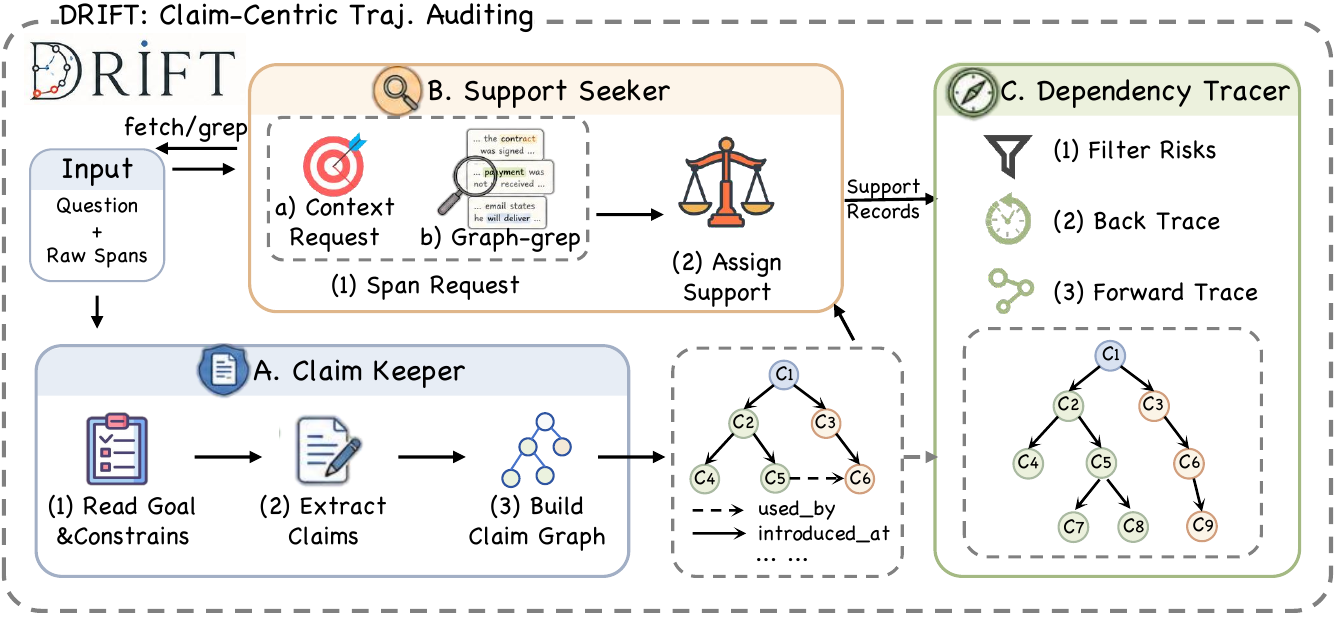}
  \caption{Overview of \textsc{DRIFT}: a claim-centric auditing workflow that builds trajectory-level claim ledgers, verifies support, and traces claim dependencies to localize first and follow-up errors.}
  \label{fig:overview}
\end{figure*}

\paragraph{Motivation and formulation.}
After trajectory collection, span segmentation, and error span annotation, we introduce \textsc{DRIFT} to localize erroneous spans in a completed trajectory. Given a task question $q$ and an ordered span sequence $T=(s_1,\ldots,s_n)$, \textsc{DRIFT} predicts a set of error spans $\hat{E}\subseteq T$:
\begin{equation}
    \hat{E}=f_{\theta}(q,T).
    \label{eq:drift-task}
\end{equation}
The external input contains only the question and raw span text; it does not use judge results, gold labels, manual notes, span types, or generated summaries. The key design choice is to audit claims rather than classify spans independently. In deep-research trajectories, many spans are exploratory: an agent searches, tests candidates, follows weak leads, and may later abandon them. Such spans are not harmful by themselves. A span becomes harmful when the agent commits to an unsupported, conflicting, or prematurely finalized claim and later reasoning treats that claim as established. Thus, \textsc{DRIFT} localizes errors through the claim-centric workflow in Figure~\ref{fig:overview}: when a claim is introduced, whether it is supported, and where it becomes consequential for the answer path.
\paragraph{A: Claim Keeper.}
\textsc{DRIFT} first performs a global pass over the full ordered trajectory to construct a claim ledger. A claim is a decision-relevant belief or commitment made by the agent, such as selecting an entity, accepting a constraint, interpreting evidence, relying on retrieval coverage, completing a computation, or deciding that no answer can be produced. For each claim, the ledger records where it is introduced, where it first becomes consequential, which later spans reuse it, its claim type, and its commitment status. We write the ledger as
\begin{equation}
    \mathcal{L}=\{c_k\}_{k=1}^{m},
    \qquad
    c_k=(a_k,i_k,b_k,U_k,\tau_k,\sigma_k).
    \label{eq:claim-ledger}
\end{equation}
Here, $a_k$ is the textual claim, $i_k$ is the span where it is introduced, $b_k$ is the first span where it becomes consequential, $U_k$ is the set of later spans that use it, $\tau_k$ denotes the claim type, and $\sigma_k$ denotes its status, such as exploratory, tentative, consequential, or finalized. This ledger separates ordinary exploration from committed reasoning: a candidate name in a query is only exploratory, whereas using that candidate as a settled premise is consequential.

\paragraph{B: Support Seeker.}
Given the claim ledger, the Support Seeker checks whether each consequential claim is supported by evidence shown in the trajectory. It assigns one of four support statuses: \textsc{direct}, \textsc{weak}, \textsc{missing}, or \textsc{conflicting}. \textsc{direct} means that the trajectory directly establishes the decisive link needed by the claim; \textsc{weak} means that related evidence exists but the decisive link is partial, implicit, snippet-based, or unchecked; \textsc{missing} means that no shown support establishes the claim; and \textsc{conflicting} means that shown evidence contradicts the claim. The Support Seeker also records the spans that provide or fail to provide support. Importantly, this stage does not output final error spans. Its role is to expose support risks for claims that may later become harmful commits.

\paragraph{C: Dependency Tracer.}
The final Dependency Tracer takes the claim ledger and support records, then determines which risky claims correspond to harmful error spans. A weak or missing support record is not sufficient by itself: the tracer must identify whether the claim is used as a commitment, propagated into later reasoning, computed from, used to give up, or finalized in the answer. \textsc{DRIFT} marks as error the spans that commit to, reuse, amplify, or finalize unsupported or conflicting consequential claims, and marks the remaining spans as non-error.

The final prediction is a set of error spans:
\begin{equation}
    \hat{E}=\{s_j \in T \mid h(s_j)=1\},
    \label{eq:final-prediction}
\end{equation}
where $h(s_j)=1$ indicates that span $s_j$ commits to, reuses, amplifies, or finalizes a harmful claim.

\section{Experiment}
\subsection{Experiment Settings}

We evaluate \textsc{TELBench} across five contemporary model families: Qwen-series, GPT-5.4~\citep{openai2026gpt54}, DeepSeek-V3.2~\citep{deepseekai2025deepseekv32pushingfrontieropen}, Claude-Sonnet-4.5, and Gemini-2.5-Pro. We also compare four diagnostic frameworks: Bare LLM, Claude Code, Codex, and \textsc{DRIFT}. Bare LLM performs simple trajectory inspection, Codex~\citep{openai2025codex} and Claude Code~\citep{anthropic2026claudecode} are adapted as general agentic auditing baselines, and \textsc{DRIFT} is our claim-centric trajectory auditing framework. All frameworks receive the same input, consisting of the question and ordered semantic spans, and are required to output the indices of error spans. Each setting is repeated three times.

\textsc{TELBench} uses verified-1K set, divided into 600 easy and 400 hard examples according to trajectory complexity and error subtlety. We report results on the full set and both difficulty splits. Evaluation metrics include first-error accuracy, macro precision, recall, F1. Error spans are treated as span-level metrics, while first-error accuracy measures detection of the earliest predicted error.

\subsection{Main Results}
\begin{table*}[tb!]
\centering
\small
\caption{Easy/hard split results. All numbers are percentages. P, R, and F1 are macro-averaged; FEA denotes first-error accuracy. Superscripts show absolute improvements over the bare baseline.}
\setlength{\tabcolsep}{4pt}
\resizebox{\textwidth}{!}{
\begin{tabular}{llcccccccccc}
\toprule
\multirow{2}{*}{Model} & \multirow{2}{*}{Method}
& \multicolumn{4}{c}{Easy}
& \multicolumn{4}{c}{Hard}
& \multicolumn{2}{c}{Overall} \\
\cmidrule(lr){3-6}
\cmidrule(lr){7-10}
\cmidrule(lr){11-12}
& & P & R & F1 & FEA
& P & R & F1 & FEA
& F1 & FEA \\
\midrule
DeepSeek-V3.2 & Bare
& 33.53 & 22.68 & 25.89 & 16.00
& 39.19 & 11.49 & 17.31 & 1.75
& 22.46 & 10.30 \\

DeepSeek-V3.2 & Codex
& 16.67\drop{16.86} & 11.88\drop{10.80} & 12.99\drop{12.90} & 8.33\drop{7.67}
& 20.95\drop{18.24} & 7.69\drop{3.80} & 10.64\drop{6.67} & 3.50\gain{1.75}
& 12.05\drop{10.41} & 6.40\drop{3.90} \\

DeepSeek-V3.2 & Claude Code
& 28.95\drop{4.58} & 20.26\drop{2.42} & 22.53\drop{3.36} & 14.67\drop{1.33}
& 33.48\drop{5.71} & 11.85\gain{0.36} & 16.61\drop{0.70} & 2.75\gain{1.00}
& 20.16\drop{2.30} & 9.90\drop{0.40} \\

\rowcolor{oursblue}
DeepSeek-V3.2 & \textsc{DRIFT} (ours)
& \textbf{65.58}\gain{32.05} & \textbf{58.86}\gain{36.18} & \textbf{57.81}\gain{31.92} & \textbf{34.50}\gain{18.50}
& \textbf{67.96}\gain{28.77} & \textbf{31.37}\gain{19.88} & \textbf{39.57}\gain{22.26} & \textbf{7.50}\gain{5.75}
& \textbf{50.51}\gain{28.05} & \textbf{23.70}\gain{13.40} \\






\midrule

GPT-5.4 & Bare
& 43.38 & 34.00 & 36.12 & 21.50
& 53.02 & 23.46 & 30.66 & 5.00
& 33.93 & 14.90 \\

GPT-5.4 & Codex
& 42.15\drop{1.23} & 35.19\gain{1.19} & 36.01\drop{0.11} & 22.00\gain{0.50}
& 52.05\drop{0.97} & 25.73\gain{2.27} & 32.19\gain{1.53} & 6.00\gain{1.00}
& 34.48\gain{0.55} & 15.60\gain{0.70} \\

GPT-5.4 & Claude Code
& 46.04\gain{2.66} & 39.78\gain{5.78} & 40.04\gain{3.92} & 24.33\gain{2.83}
& 54.52\gain{1.50} & 27.76\gain{4.30} & 34.08\gain{3.42} & \textbf{7.25}\gain{2.25}
& 37.66\gain{3.73} & 17.50\gain{2.60} \\

\rowcolor{oursblue}
GPT-5.4 & \textsc{DRIFT} (ours)
& \textbf{64.19}\gain{20.81} & \textbf{63.33}\gain{29.33} & \textbf{58.45}\gain{22.33} & \textbf{29.83}\gain{8.33}
& \textbf{69.14}\gain{16.12} & \textbf{35.59}\gain{12.13} & \textbf{43.51}\gain{12.85} & \textbf{7.25}\gain{2.25}
& \textbf{52.48}\gain{18.55} & \textbf{20.80}\gain{5.90} \\

\midrule
Claude-Sonnet-4.6 & Bare
& 29.78 & 21.99 & 24.01 & 15.67
& 38.56 & 12.97 & 18.71 & 4.75
& 21.89 & 11.30 \\

Claude-Sonnet-4.6 & Codex
& 32.12\gain{2.34} & 24.76\gain{2.77} & 26.35\gain{2.34} & 16.67\gain{1.00}
& 42.76\gain{4.20} & 15.16\gain{2.19} & 21.32\gain{2.61} & 4.75
& 24.34\gain{2.45} & 11.90\gain{0.60} \\

Claude-Sonnet-4.6 & Claude Code
& 40.22\gain{10.44} & 30.06\gain{8.07} & 32.71\gain{8.70} & 20.83\gain{5.16}
& 49.81\gain{11.25} & 18.09\gain{5.12} & 25.35\gain{6.64} & 6.00\gain{1.25}
& 29.77\gain{7.88} & 14.90\gain{3.60} \\

\rowcolor{oursblue}
Claude-Sonnet-4.6 & DRIFT (ours)
& \textbf{63.00}\gain{33.22} & \textbf{67.31}\gain{45.32} & \textbf{60.00}\gain{35.99} & \textbf{32.17}\gain{16.50}
& \textbf{68.39}\gain{29.83} & \textbf{41.16}\gain{28.19} & \textbf{47.28}\gain{28.57} & \textbf{12.00}\gain{7.25}
& \textbf{54.91}\gain{33.02} & \textbf{24.10}\gain{12.80} \\

\midrule
Gemini-2.5-Pro & Bare
& 38.59 & 33.12 & 33.39 & 20.50
& 44.61 & 22.55 & 27.44 & 8.50
& 31.01 & 15.70 \\

Gemini-2.5-Pro & Codex
& 38.83\gain{0.24} & 39.19\gain{6.07} & 36.16\gain{2.77} & 19.17\drop{1.33}
& 48.88\gain{4.27} & 27.90\gain{0.46} & 33.23\gain{5.79} & \textbf{10.50}\gain{2.00}
& 34.99\gain{3.98} & 15.70\\

Gemini-2.5-Pro & Claude Code
& 34.86\drop{3.73} & 33.47\gain{0.35} & 31.48\drop{1.91} & 18.00\drop{2.50}
& 40.03\drop{4.58} & 20.56\drop{1.99} & 25.36\drop{2.08} & 8.50
& 29.03\drop{1.98} & 14.20\drop{1.5} \\

\rowcolor{oursblue}
Gemini-2.5-Pro & DRIFT (ours)
& \textbf{56.62}\gain{18.03} & \textbf{58.06}\gain{24.94} & \textbf{52.94}\gain{19.55} & \textbf{27.17}\gain{6.67}
& \textbf{63.81}\gain{19.20} & \textbf{35.02}\gain{12.47} & \textbf{41.62}\gain{14.18} & 9.00\gain{0.50}
& \textbf{48.41}\gain{17.40} & \textbf{19.90}\gain{4.20} \\

\bottomrule
\end{tabular}
}

\label{tab:difficulty-split}
\end{table*}

\begin{figure*}[tb!]
  \centering
  \includegraphics[width=1\textwidth]{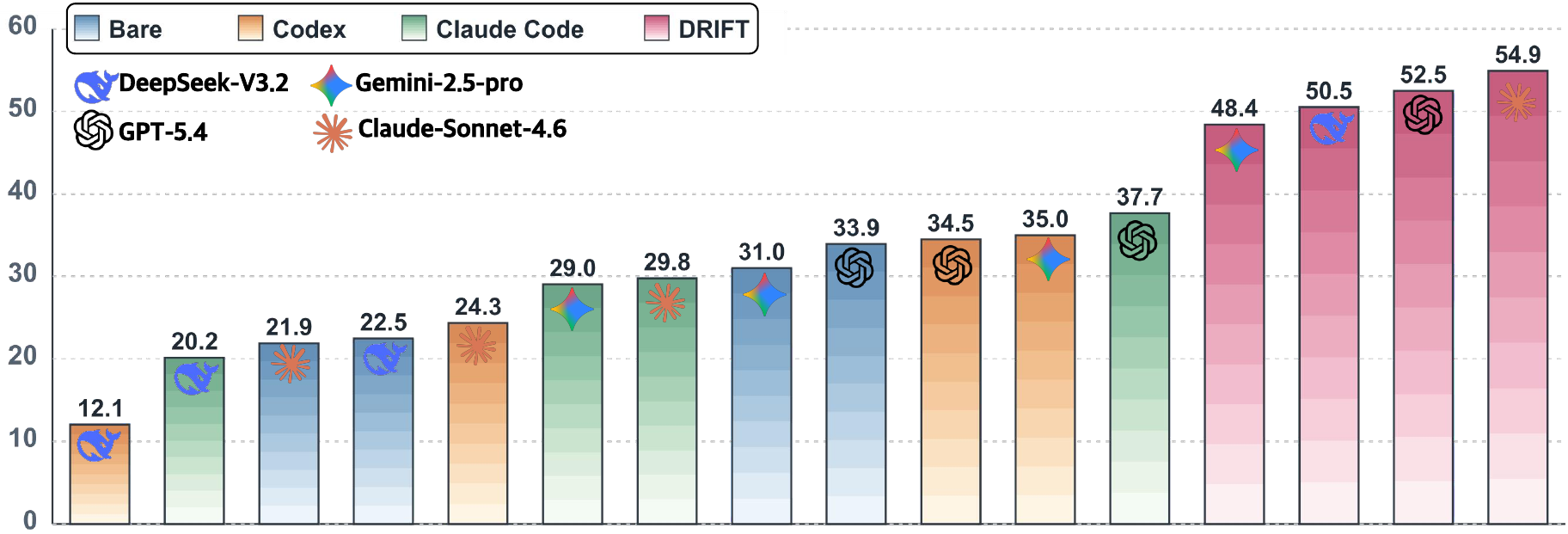}
  \caption{\textbf{Overall macro-F1} on \textsc{TELBench}.}
  \label{fig:performance}
\end{figure*}

\paragraph{\textsc{DRIFT} outperforms generic auditing frameworks.}
Table~\ref{tab:difficulty-split} and Figure~\ref{fig:performance} shows that \textsc{DRIFT} achieves the best overall F1 across all backbone models, outperforming both bare full trajectory prompting and general agentic auditing frameworks such as Codex and Claude Code. The comparison indicates that simply wrapping an LLM in a more complex agentic workflow is not sufficient for reliable trajectory diagnosis: Codex and Claude Code bring inconsistent gains and can even degrade performance for some backbones. In contrast, \textsc{DRIFT} improves both precision and recall across easy and hard splits, suggesting that its gains do not come from over-predicting suspicious spans. Instead, the claim ledger helps track consequential commitments, support seeking checks whether those commitments are grounded in trajectory evidence, and dependency tracing filters out normal exploration, tentative hypotheses, and harmless noise. This claim-centric bias makes \textsc{DRIFT} more effective at separating harmful error spans from surrounding non-error behavior.

\paragraph{First-error localization remains difficult.}
Although \textsc{DRIFT} substantially improves span-level F1, first-error accuracy remains much lower, especially on the hard split. This gap shows that identifying some erroneous regions and identifying where the error first appears are distinct diagnostic abilities. Current auditors can often detect that a trajectory has become unreliable, but still struggle to pinpoint the earliest annotated error span among long sequences of search, verification, and intermediate reasoning. \textsc{TELBench} therefore evaluates not only aggregate error span localization, but also the stricter temporal localization ability required to diagnose where first goes wrong.

\paragraph{Scaling alone is insufficient.}
Figure~\ref{fig:further-analysis}(a) further shows that increasing model scale does not monotonically improve trajectory diagnosis. Across Qwen variants, larger models do not consistently achieve better macro F1 or first-error accuracy, and the hard split remains challenging for all scales. This suggests that the bottleneck is not only backbone capacity, but also the absence of a diagnostic structure tailored to long, noisy agent trajectories.

\section{Further Analysis}
\begin{figure}[t]
\centering
\begin{minipage}{0.48\textwidth}
  \centering
  \includegraphics[width=\linewidth]{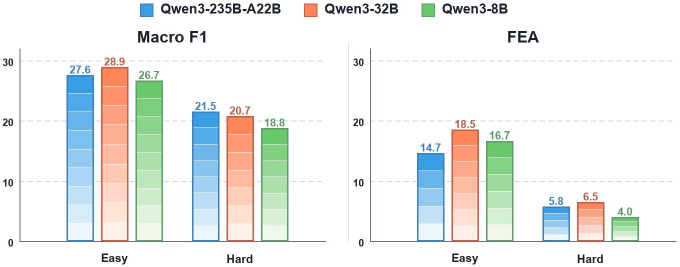}
  \vspace{-6pt}
  \caption*{\small (a) Model-scale sensitivity.}
\end{minipage}
\hfill
\begin{minipage}{0.48\textwidth}
  \centering
  \includegraphics[width=\linewidth]{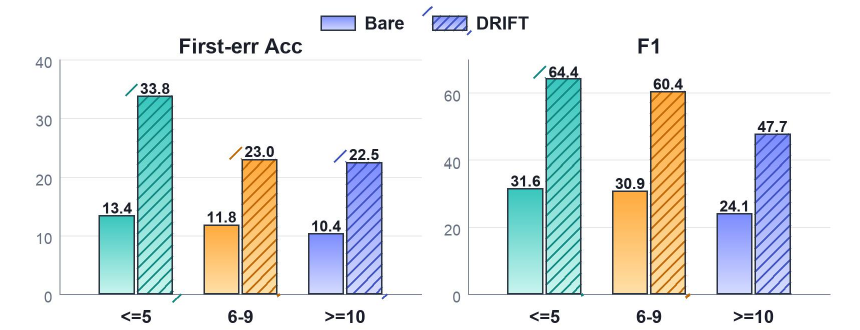}
  \vspace{-6pt}
  \caption*{\small (b) Span-complexity sensitivity.}
\end{minipage}

\vspace{4pt}

\begin{minipage}{0.48\textwidth}
  \centering
  \includegraphics[width=\linewidth]{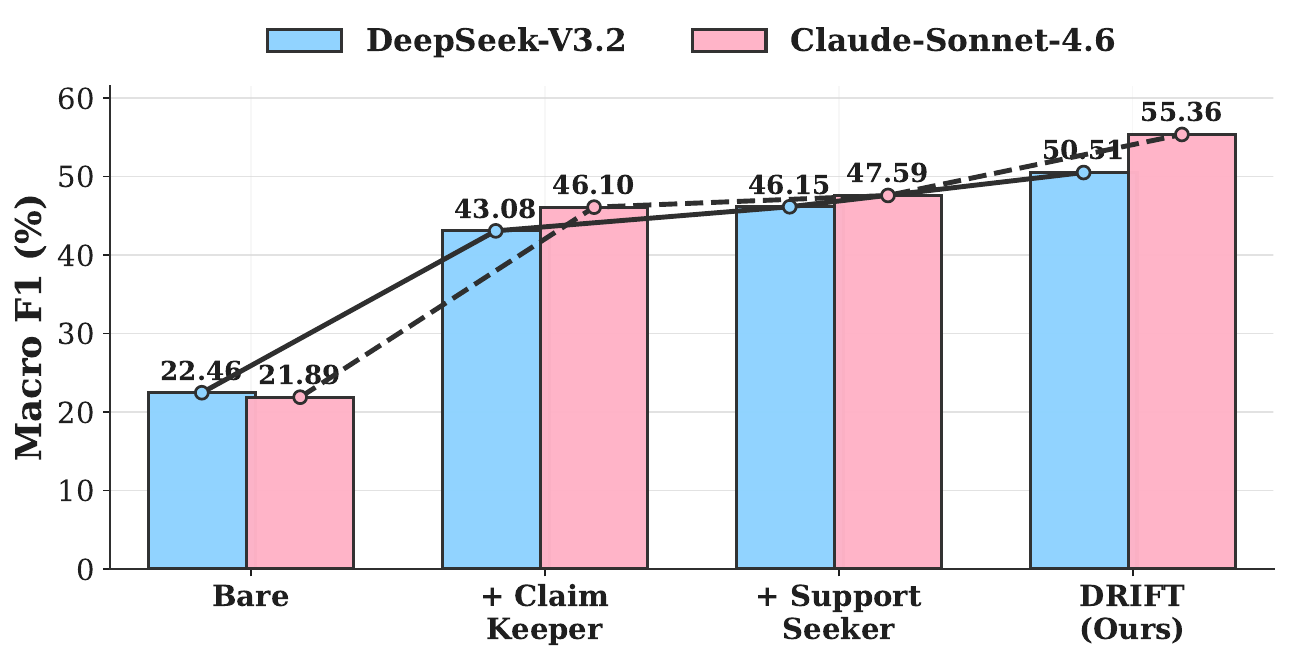}
  \vspace{-6pt}
  \caption*{\small (c) Module ablation.}
\end{minipage}
\hfill
\begin{minipage}{0.48\textwidth}
  \centering
  \includegraphics[width=\linewidth]{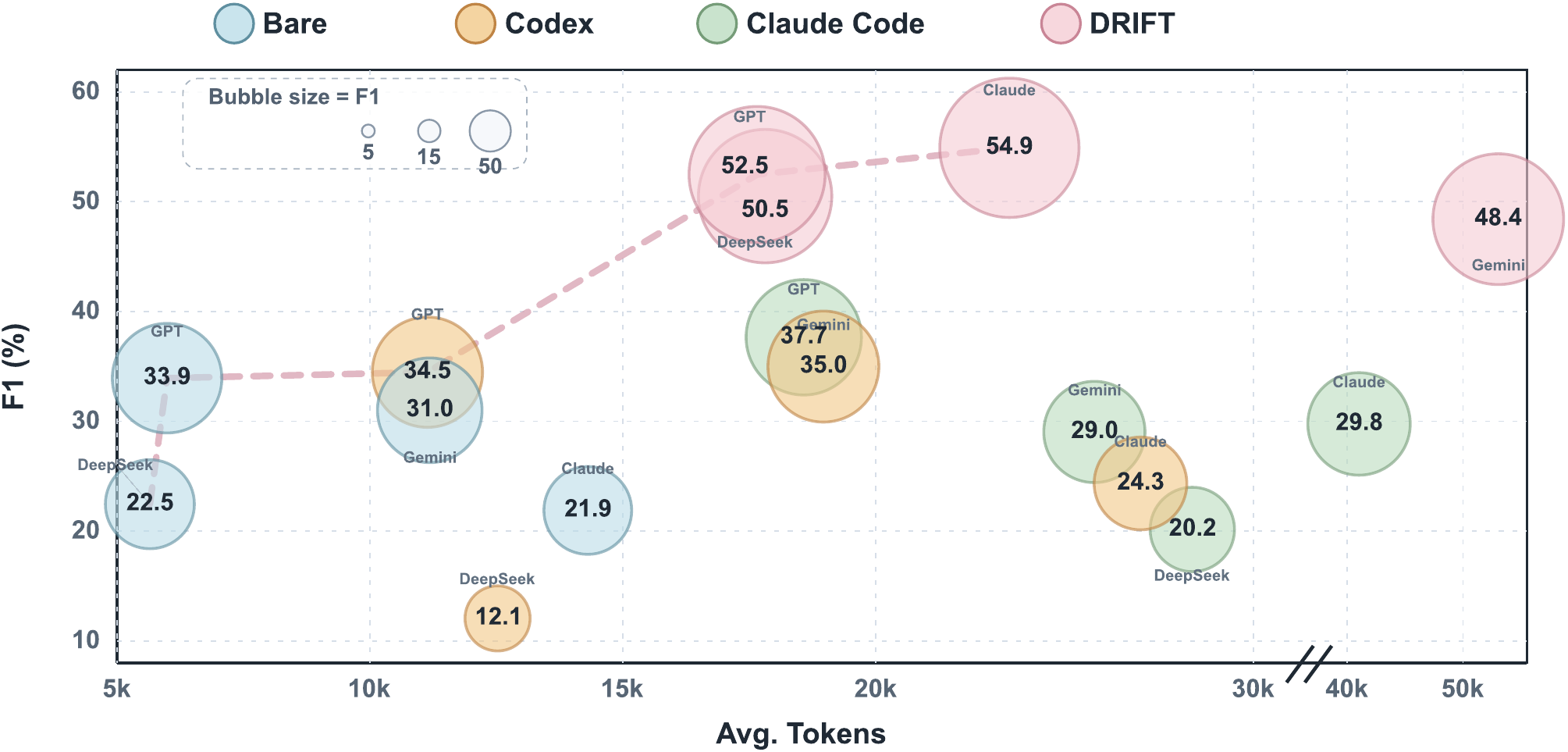}
  \vspace{-6pt}
  \caption*{\small (d) Efficiency-performance trade-off.}
\end{minipage}

\vspace{-4pt}
\caption{
Further analysis of \textsc{DRIFT}. We examine robustness across model scale and span complexity, then verify that the gains come from the proposed modules and remain competitive under token cost.
}
\label{fig:further-analysis}
\vspace{-8pt}
\end{figure}

\paragraph{Sensitivity to Span Complexity.}
Figure~\ref{fig:further-analysis}(b), as span complexity increases, both Bare and DRIFT degrade, showing that longer trajectories make error span localization harder. DRIFT consistently outperforms Bare across all span buckets, suggesting that structured trajectory auditing better preserves localization ability under longer semantic contexts. The gap is especially visible in high-span trajectories, where single-pass reading is more likely to miss early or distributed errors.
\paragraph{Ablation of modules.}

Figure~\ref{fig:further-analysis}(c) compares four variants: bare prediction with the full trajectory, A (Claim Keeper), A+B (support checking), and the full \textsc{DRIFT} pipeline with dependency tracing. Performance improves steadily as modules are added. The largest gain comes from claim-level auditing, while support checking and dependency tracing further improve evidence grounding and span localization. This shows that \textsc{DRIFT}'s gains arise from the complementary effects of claim tracking, support auditing, and dependency-based diagnosis.

\paragraph{Efficiency Analysis.}

Overall, in Figure~\ref{fig:further-analysis}(d), DRIFT achieves a favorable efficiency-performance trade-off and mostly lies on the Pareto frontier, indicating that it improves F1 without requiring disproportionate token overhead. The only notable exception is Gemini, whose DRIFT run incurs substantially higher cost because more than half of its tokens are spent on thinking, leading to a much larger average token budget despite competitive performance.
\paragraph{Error-type coverage: Can DRIFT detect all kinds of error?}

\begin{figure}[t]
  \centering
  \includegraphics[width=0.6\textwidth]{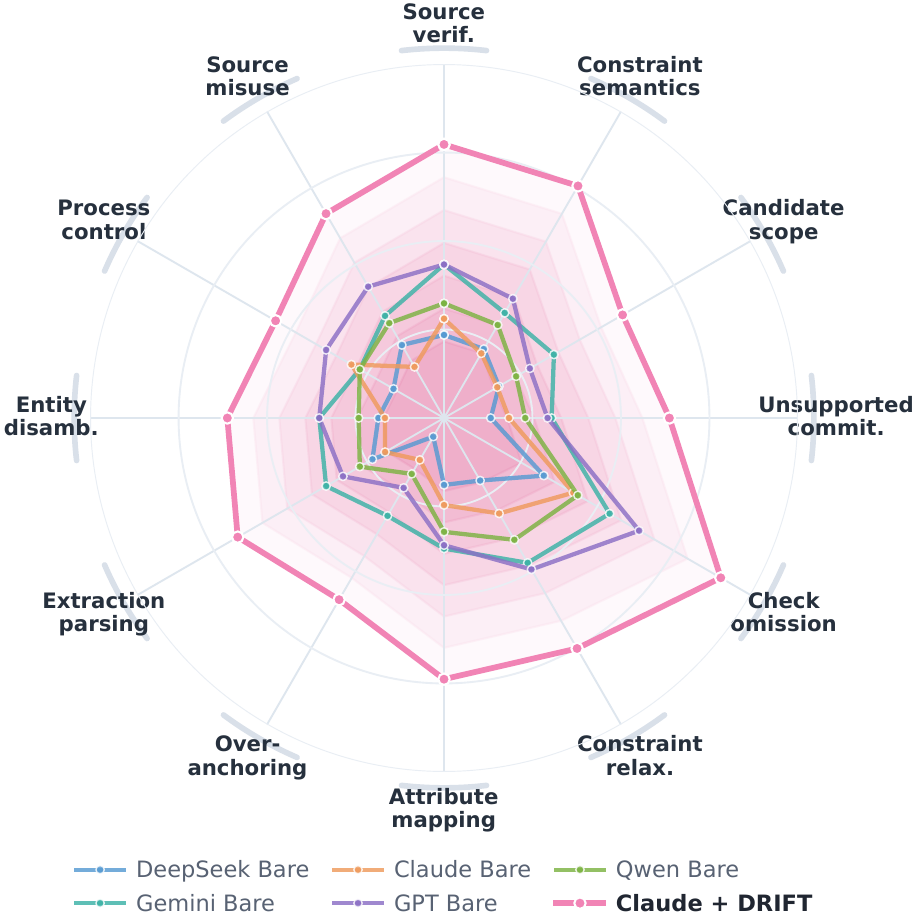}
  \vspace{-8pt}
  \caption{
  Span-level recall across frequent error types. \textsc{DRIFT} improves coverage especially on evidence- and constraint-related failures.
  }
  \label{fig:error-type-radar}
  \vspace{-8pt}
\end{figure}

Figure~\ref{fig:error-type-radar} reports span-level recall on the 12 most frequent error types. Bare models show uneven coverage: they recover some explicit errors, but struggle with failures that require verifying whether a claim is sufficiently supported, such as source verification, constraint semantics, unsupported commitments, and omitted constraint checks. \textsc{DRIFT} improves recall consistently across nearly all categories, with the largest gains on evidence- and constraint-related errors. This aligns with its claim-centric design: the claim ledger tracks consequential commitments, support seeking checks their evidential basis, and dependency tracing marks the spans where unsupported claims affect the answer path. The result suggests that \textsc{DRIFT} improves not only overall localization performance, but also robustness across diverse high-frequency failure modes.

\section{Conclusion}
We study deep-research agent reliability beyond final answer correctness by formulating span-level error localization over semantic trajectories. We introduce \textsc{TELBench}, a benchmark built from real agent runs that tests whether models can distinguish harmful error spans from benign trajectory behavior. This setting captures the central difficulty of deep research: errors often emerge when weakly supported claims are repeatedly reused as evidence. We further propose \textsc{DRIFT}, a claim-centric auditing framework that checks whether agent claims are supported by trajectory evidence and marks spans where unsupported or conflicting claims affect the answer path. Experiments show that \textsc{DRIFT} outperforms bare prompting and generic agentic auditors, while scaling alone is insufficient and first-error localization remains challenging. Our results highlight the need to evaluate deep-research agents through process level reliability, rather than final outcomes alone.

\bibliographystyle{unsrtnat}
\bibliography{example_paper}
\appendix
\section*{Appendix}
\section{Annotation Guidelines and Annotator Information}
\label{app:annotation-guidelines}
\begin{figure*}[htb!]
  \centering
  \includegraphics[width=0.6\textwidth]{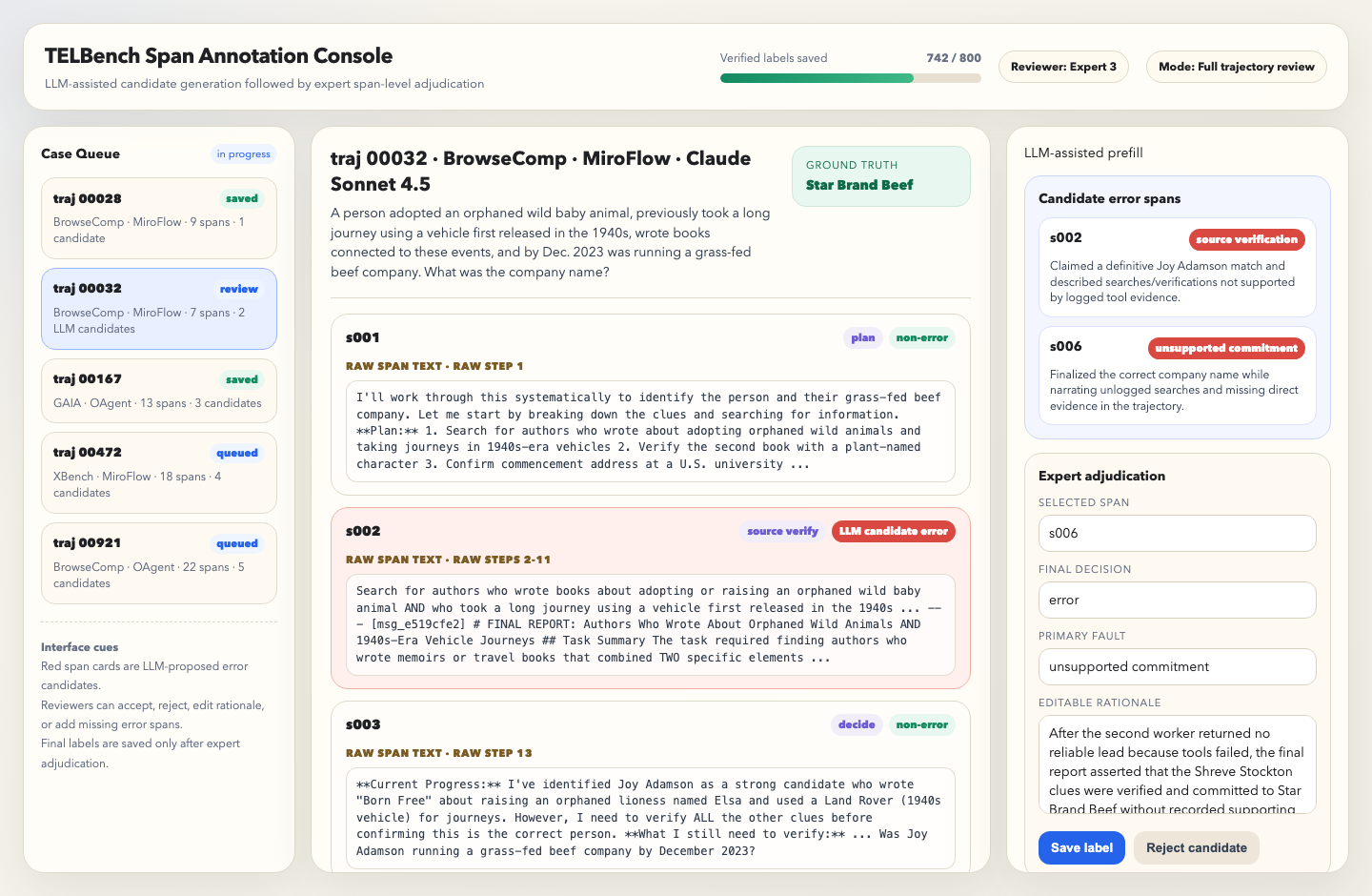}
  \caption{Annotation interface for expert span-level adjudication. The console shows the ordered semantic spans, LLM-assisted candidate errors, editable rationales, and final expert decisions.}
  \label{fig:annotation-ui}
\end{figure*}
\paragraph{Annotation interface.}
Figure~\ref{fig:annotation-ui} shows the annotation console used by expert annotators. The interface presents the case metadata, task question, ground-truth answer, ordered semantic spans, and LLM-assisted candidate error spans with proposed rationales. Candidate spans are highlighted in red, while non-error spans and span-stage cues are shown separately to help annotators distinguish harmful errors from normal exploration. Annotators can inspect the full trajectory, select a span, accept or reject LLM-proposed candidates, edit the primary fault and rationale, and save the final adjudicated label. Labels are saved only after expert review, rather than directly copied from the LLM-assisted prefill.


\section{Detailed Experiment Setting}

\paragraph{Framework and tooling setup.}
To make cross-framework comparisons interpretable, we control external factors that can substantially shift agent behavior, especially retrieval and reading tools. We use Serper as the unified search interface and Jina as the unified reading interface across the agent frameworks, reducing confounding effects from different search APIs and page-reading implementations. For non-retrieval tools, such as code execution and audio/image/video understanding, we keep each framework's native configuration because these tools are tightly coupled with framework-specific wrappers, callback formats, and error-handling policies. Forcing a fully unified toolchain would introduce additional implementation bias. In our runs, MiroFlow uses claude-3-7-sonnet-20250219 for image and video understanding, E2B Sandbox for code execution, gpt-4o-audio-preview for audio understanding, and claude-sonnet-4-5-20250929-thinking as the primary reasoning model. OAgent uses Serper for search and Jina for reading.

\section{Detailed Error Analysis for Deep-research Agent Systems.}
Unless otherwise stated, our mechanism analysis is conducted on the full annotated corpus of 2,790 trajectories; the Verified-1K subset is used only for benchmark evaluation.
\subsection{Basic Analysis}
\label{app:basic-analysis}
\begin{figure*}[tb!]
  \centering
  \includegraphics[width=1\textwidth]{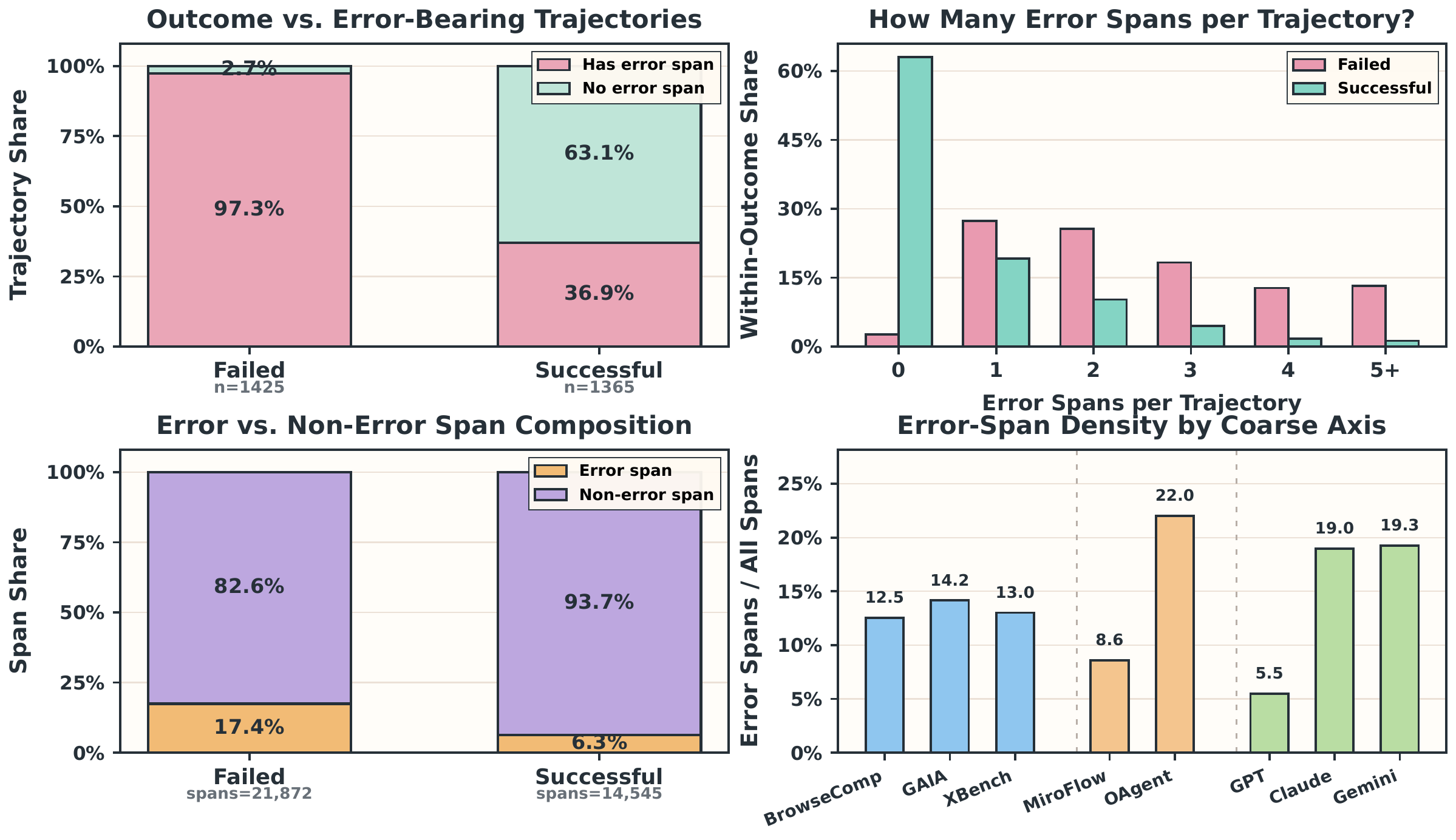}
  \vspace{-15pt}
  \caption{
  Basic error-burden statistics of annotated trajectories.
  We compare final failed and successful trajectories by whether they contain any annotated error span, the number of error spans per trajectory, the composition of error versus non-error spans, and the overall error spans density across benchmarks, frameworks, and model families.
  }
  \label{fig:error_burden_basics}
  \vspace{-10pt}
\end{figure*}

\paragraph{Error burden.}
Before analyzing specific fault mechanisms, we first summarize the basic scale of annotated process errors.
Figure~\ref{fig:error_burden_basics} compares failed and successful trajectories by whether they contain any annotated error span, how many error spans appear in each trajectory, how error and non-error spans are composed at the span level, and how dense error spans are across coarse data and system axes.
This provides a sanity check for our span-level annotation: process errors are closely related to final answer failure, but they are not equivalent to it.

\begin{figure*}[tb!]
  \centering
  \includegraphics[width=0.95\textwidth]{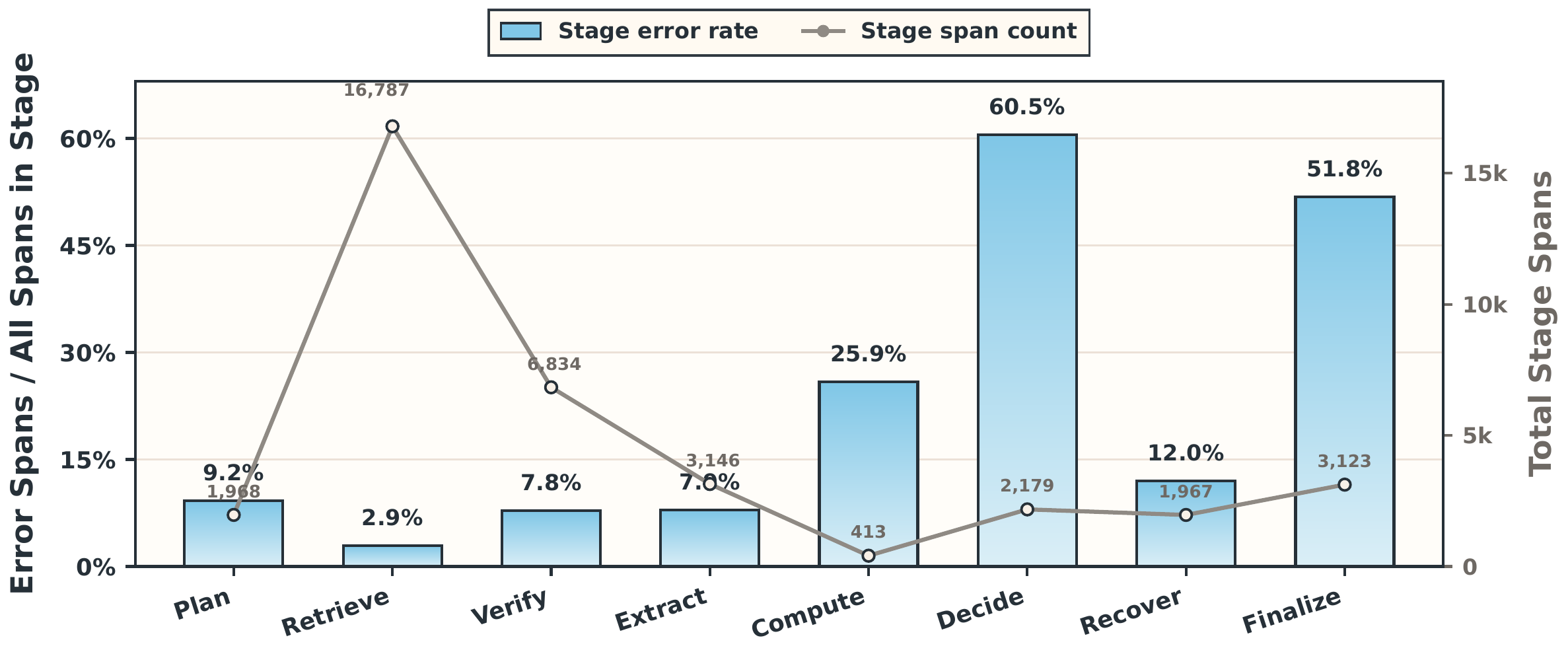}
  \vspace{-10pt}
  \caption{
  Stage-normalized error rates across operation stages.
  Bars show the percentage of spans in each stage that are annotated as errors, while the gray line reports the total number of spans assigned to that stage.
  This normalization separates stages that are common in trajectories from stages that are intrinsically more error-prone.
  }
  \label{fig:stage_normalized_error_rate}
  \vspace{-10pt}
\end{figure*}

final answer failure is strongly associated with process errors: 97.3\% of failed trajectories contain at least one annotated error span.
However, process errors are not identical to final failure.
Among successful trajectories, 36.9\% still contain at least one error span, showing that agents can recover from local mistakes or reach the correct answer despite unsupported intermediate commitments.
Failed trajectories are also much more likely to contain multiple error spans, including cases with five or more annotated error spans, while successful trajectories are dominated by zero- or one-error cases.
At the span level, failed trajectories have a much higher error span share than successful trajectories: 17.4\% of spans in failed trajectories are annotated as errors, compared with 6.3\% in successful trajectories.
This shows that final failure is associated not only with whether an error appears, but also with how much of the trajectory becomes error-bearing.
At the same time, most spans remain non-error spans even in failed trajectories, reinforcing that our annotation targets specific harmful commitments rather than broadly labeling entire failed traces as erroneous.
\paragraph{Stage-normalized error risk.}
\label{app:stage-risk}
The stage--fault heatmap in the main text shows where annotated errors occur, but raw counts can be affected by how often a stage appears.
For example, retrieval spans are frequent in long research trajectories, so a large number of retrieval-stage errors does not necessarily mean that retrieval is the riskiest stage.
To separate stage prevalence from stage risk, Figure~\ref{fig:stage_normalized_error_rate} reports the normalized error rate for each operation stage, computed as the number of error spans in that stage divided by the total number of spans assigned to that stage.
The gray line shows the denominator, i.e., the total number of spans in each stage.

\begin{figure*}[tb!]
  \centering
  \includegraphics[width=1\textwidth]{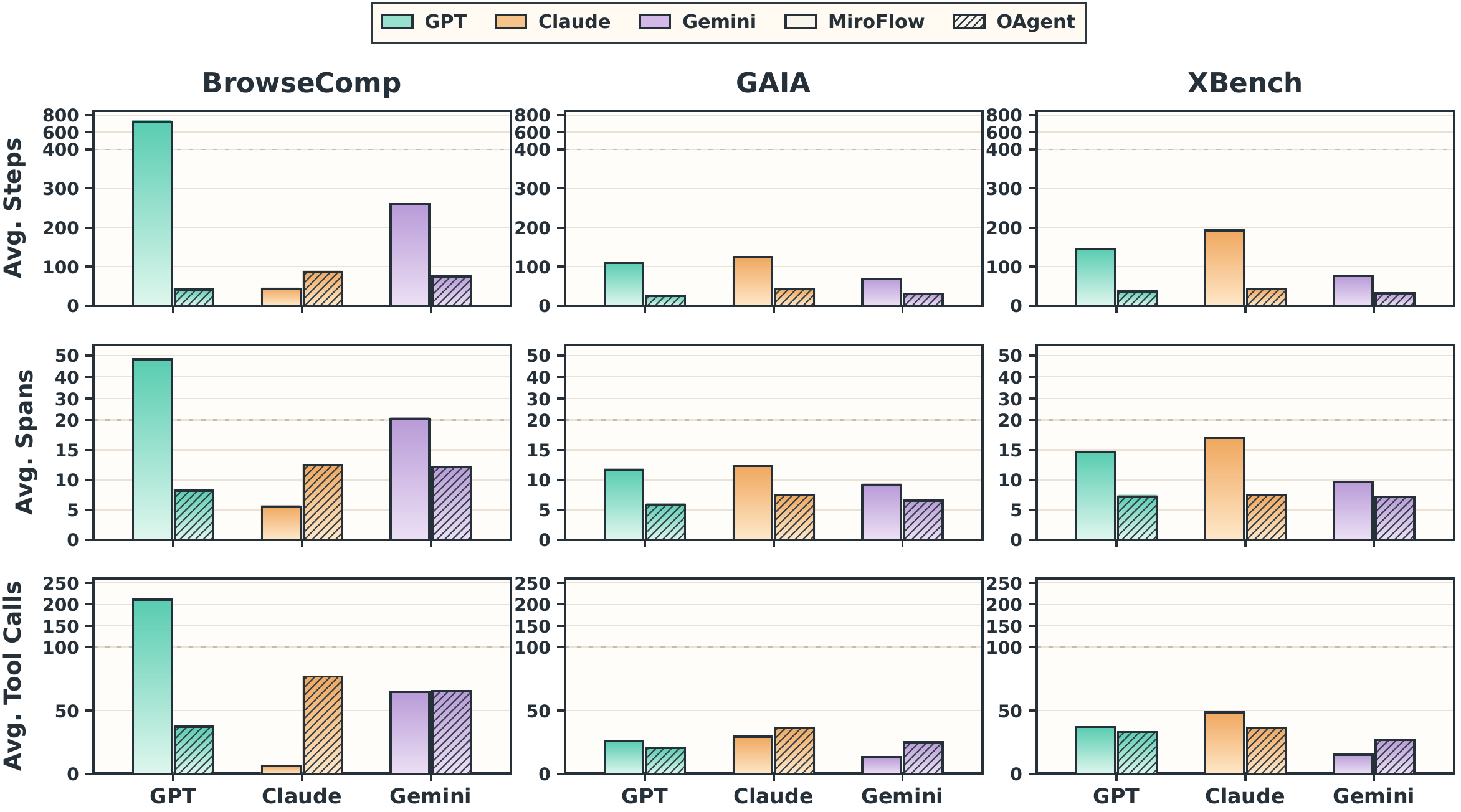}
  \vspace{-15pt}
  \caption{
  Effort profiles across benchmarks, frameworks, and models.
  Average trajectory steps, annotated spans, and tool calls are reported for each benchmark--model--framework combination.
  Colors denote model families, while hatching distinguishes frameworks.
  The y-axes are piecewise-compressed above 400 steps, 20 spans, and 100 tool calls to preserve visibility of lower-effort settings.
  }
  \label{fig:effort_profiles}
  \vspace{-10pt}
\end{figure*}

Figure~\ref{fig:stage_normalized_error_rate} shows that retrieval dominates the trajectory volume but has the lowest normalized error rate.
Only 2.9\% of retrieval spans are annotated as errors, despite retrieval accounting for the largest number of spans.
In contrast, decision-making and finalization are much more error-prone, with normalized error rates of 60.5\% and 51.8\%, respectively.
Compute spans also have a relatively high error rate, although they occur much less frequently.
This suggests that many failures are not caused by search activity itself, but by how agents commit to, verify, aggregate, or finalize the information gathered during earlier stages.
\paragraph{Effort profiles.}
We next examine how much trajectory effort different systems spend before final prediction.
Figure~\ref{fig:effort_profiles} reports average trajectory steps, annotated spans, and tool calls for each benchmark--model--framework combination.
This analysis is orthogonal to accuracy: it describes the execution behavior of the agent systems rather than whether their final answers are correct.

Across benchmarks, MiroFlow generally produces longer trajectories with more intermediate spans, especially for GPT on BrowseComp, suggesting a more expansive decomposition and search process.
In contrast, OAgent tends to maintain shorter trajectories, although its tool usage can still be high for some model--benchmark pairs.
This indicates that fewer reasoning steps do not necessarily imply fewer external actions.
These effort profiles help separate performance from execution behavior: models and frameworks may reach similar task outcomes through very different amounts of planning, evidence gathering, and tool interaction.

\subsection{Operation Stage Taxonomy}
\label{app:taxonomy}
We annotate every trajectory span with one operation stage that describes the functional role of the span in the agent process. This label is independent of correctness: both error and non-error spans receive a stage label. Table~\ref{tab:operation-stages} lists eight stages, covering the main phases of long-form agent behavior from decomposing the task and searching for evidence to verifying sources, extracting information, making decisions, recovering from conflicts, and producing the final answer.

The stage taxonomy lets us analyze where errors occur relative to the agent's process, rather than only asking whether the final answer is correct. Because all spans receive a stage label, the stage distribution also provides a denominator for error-rate analysis. This supports comparisons such as whether one framework spends more trajectory mass in retrieval, whether a benchmark induces more source verification failures, or whether errors tend to appear earlier in decision-making but become finalized later in the trajectory.

\begin{table}[t]
\centering
\small
\begin{tabular}{p{0.22\linewidth} p{0.68\linewidth}}
\toprule
Stage & Definition \\
\midrule
Plan & Task decomposition, goal framing, subgoal design, or deciding what information needs to be collected. \\
Retrieve & Search, browsing, query construction, candidate enumeration, or opening potentially relevant sources. \\
Source Verify & Checking whether a source is reliable, relevant, accessible, or whether the evidence supports a claim. \\
Extract & Extracting fields, relations, dates, values, names, or other structured information from evidence. \\
Compute & Calculation, counting, aggregation, unit conversion, numerical comparison, or metric computation. \\
Decide & Comparing candidates, excluding alternatives, selecting a candidate, or committing to an answer before final submission. \\
Reflect Recover & Self-checking, recognizing conflict, revising a route, rolling back a candidate, or recovering from a failed path. \\
Finalize & Producing the final answer, final report, boxed response, or submission-facing summary. \\
\bottomrule
\end{tabular}

\caption{Operation stage taxonomy used for trajectory span annotation. Each span is assigned exactly one stage, regardless of whether it is an error span. The stage label describes the functional role of the span in the agent trajectory rather than its correctness.}
\label{tab:operation-stages}

\end{table}

\begin{table}[t]
\centering
\small

\begin{tabular}{p{0.24\linewidth} p{0.66\linewidth}}
\toprule
Fault Family & Primary Faults \\
\midrule
Constraint Handling & Constraint Semantics Error; Constraint Check Omission; Constraint Relaxation; Answer Format Error. \\
Search and Retrieval & Goal Drift; Candidate Scope Error; Retrieval Query Error. \\
Evidence Grounding & Source Verification Error; Source Misuse Error; Unsupported Commitment. \\
Entity Mapping & Entity Disambiguation Error; Entity Attribute Mapping Error; Memory Context Error. \\
Information Processing & Extraction Parsing Error; Calculation Error; Aggregation Metric Error. \\
Process Control & Overanchoring Error; Process Control Error. \\
\bottomrule
\end{tabular}
\caption{Error taxonomy used for error span annotation. Each error span is assigned exactly one primary fault, which is grouped into one of six broader fault families. Non-error spans do not receive a fault label.}
\label{tab:error-taxonomy}
\end{table}

\subsection{Error Fault Taxonomy}
\subsubsection{Construction}
Our error taxonomy is not manually enumerated in advance. Instead, it is induced from the completed span-level error annotations. The construction process consists of three rounds: error-rationale generation, candidate type induction, and final taxonomy normalization with back-labeling.

In the first round, we generate error rationales for annotated error spans. We use three frontier LLMs as independent annotators. For each error span, the model is given the question, trajectory context, span content, and the span-level error judgment, and is asked to produce a free-form explanation of why the span is erroneous. At this stage, the model is not asked to choose from any predefined taxonomy. Instead, it describes the underlying failure mechanism in natural language, such as constraint misinterpretation, candidate-scope drift, incorrect entity binding, unverified evidence, metric or calculation errors, or premature commitment. We then extract short error keywords, or error-reason keys, from these rationales. After cleaning, deduplication, and filtering, we obtain 4,631 error-reason keys as the input to the next round.

In the second round, we induce candidate error types in a bottom-up manner. To avoid topic drift from a single long-context clustering step, and to prevent a few frequent patterns from dominating the entire taxonomy, we use a hierarchical map-reduce procedure. We first randomly shuffle the 4,631 keys and split them into 58 chunks with a chunk size of 80, with the final chunk containing the remaining keys. In the map stage, each chunk independently produces 10 local error types. We then perform three levels of reduce. Reduce-1 merges every 10 chunks into one mid-level taxonomy; the 58 chunks are grouped into six groups of sizes 10, 10, 10, 10, 10, and 8, and each group outputs approximately 18--25 candidate types. Reduce-2 further merges the six mid-level taxonomies into two higher-level taxonomies, each retaining approximately 14--20 types. Reduce-3 merges these two taxonomies into taxonomy v0, controlled to contain 12--18 candidate types. This hierarchical procedure preserves local diversity while preventing the final taxonomy from collapsing into overly coarse categories.

In the third round, we normalize the candidate taxonomy, calibrate category boundaries, and validate it by back-labeling. We manually inspect taxonomy v0 with a focus on three issues: synonymous duplicates, overlapping boundaries, and overly narrow long-tail categories. Semantically similar types are merged; for example, ``unsupported claim,'' ``unverified submission,'' and ``final conclusion without support'' are unified under Unsupported Commitment. In contrast, superficially similar but mechanistically different types are separated; for example, failing to verify whether evidence exists and using an incorrect or inapplicable source correspond to Source Verification Error and Source Misuse Error, respectively. We then write definitions, inclusion criteria, and exclusion criteria for each final type, and organize the 18 primary faults into six broader fault families. Finally, we map the finalized taxonomy back to all error spans: each error span receives exactly one primary fault, while non-error spans receive no fault label. After back-labeling, we inspect category coverage, long-tail distribution, commonly confused pairs, and random samples, and revise a small number of boundary cases when needed.

This process serves two goals. First, the taxonomy is grounded in localized error span rationales from real trajectories, rather than being inferred from final answer correctness. Second, through hierarchical induction and manual boundary calibration, the taxonomy yields a stable category structure for analyzing error patterns across frameworks, models, and benchmarks.

\subsubsection{Analysis}

For each span annotated as erroneous, we assign exactly one primary fault label. While the operation stage captures what the agent was doing at that point in the trajectory, the primary fault captures why the span is erroneous. The label therefore describes the underlying failure mechanism rather than the surface form or position of the span. Non-error spans receive no fault label.

Table~\ref{tab:error-taxonomy} summarizes 18 primary faults organized into six broader fault families: Constraint Handling, Search and Retrieval, Evidence Grounding, Entity Mapping, Information Processing, and Process Control. This two-level design balances granularity and comparability. Primary faults support fine-grained diagnosis of concrete failure modes, such as unsupported commitments, source verification failures, candidate scope errors, or constraint misinterpretations. Fault families provide a more stable abstraction for comparing error patterns across frameworks, models, and benchmarks.

In error spans analysis, we use the two annotations jointly. The stage label tells us where in the agent process an error occurs, such as retrieval, verification, decision-making, or finalization. The fault label tells us what kind of mechanism caused the error, such as misread constraints, unsupported evidence, wrong entity mapping, or flawed computation. This separation allows us to distinguish, for example, a retrieval-stage error caused by a poor query from a retrieval-stage error caused by drifting to the wrong candidate set, and to compare whether different systems fail at similar stages for different reasons.

\section{Token Consumption}

\begin{table}[tb!]
\centering
\caption{Token consumption on the main benchmark. Prompt and completion tokens are summed over all trajectories. Avg. denotes average total tokens per trajectory.}
\scriptsize
\setlength{\tabcolsep}{3pt}
\renewcommand{\arraystretch}{1.06}
\resizebox{0.6\columnwidth}{!}{%
\begin{tabular}{llrrr}
\toprule
\textbf{Model} & \textbf{Method} & \textbf{Prompt} & \textbf{Completion} & \textbf{Avg.} \\
\midrule
DeepSeek-V3.2 & Bare & 5,569,782 & 79,480 & 5,649 \\
DeepSeek-V3.2 & Codex & 12,327,974 & 194,502 & 12,522 \\
DeepSeek-V3.2 & Claude Code & 27,485,517 & 620,830 & 28,106 \\
\rowcolor{oursblue}
DeepSeek-V3.2 & \textsc{DRIFT} & 16,950,426 & 861,475 & 17,812 \\
\midrule
GPT-5.4 & Bare & 5,912,241 & 76,063 & 5,988 \\
GPT-5.4 & Codex & 11,052,241 & 83,816 & 11,136 \\
GPT-5.4 & Claude Code & 17,998,009 & 578,129 & 18,576 \\
\rowcolor{oursblue}
GPT-5.4 & \textsc{DRIFT} & 16,352,335 & 1,300,775 & 17,653 \\
\midrule
Gemini-2.5-Pro & Bare & 8,106,948 & 3,074,795 & 11,182 \\
Gemini-2.5-Pro & Codex & 13,806,700 & 5,158,793 & 18,965 \\
Gemini-2.5-Pro & Claude Code & 19,154,184 & 5,918,027 & 25,072 \\
\rowcolor{oursblue}
Gemini-2.5-Pro & \textsc{DRIFT} & 18,672,368 & 34,370,710 & 53,043 \\
\midrule
Claude-Sonnet-4.6 & Bare & 14,153,744 & 110,533 & 14,307 \\
Claude-Sonnet-4.6 & Codex & 26,347,527 & 155,221 & 26,503 \\
Claude-Sonnet-4.6 & Claude Code & 40,433,306 & 602,831 & 41,036 \\
\rowcolor{oursblue}
Claude-Sonnet-4.6 & \textsc{DRIFT} & 20,302,960 & 2,339,670 & 22,643 \\
\bottomrule
\end{tabular}%
}

\label{tab:token-consumption}
\end{table}

As shown in Table~\ref{tab:token-consumption}, different agent frameworks introduce substantially different token overheads across the same benchmark. The table reports the total prompt and completion tokens accumulated over all trajectories, together with the average total tokens per trajectory.

\section{Ablation Study}
\subsection{Full ablation trends.}
\label{app:ablation_module}
\begin{figure*}[tb!]
  \centering
  
  \includegraphics[width=1\textwidth]{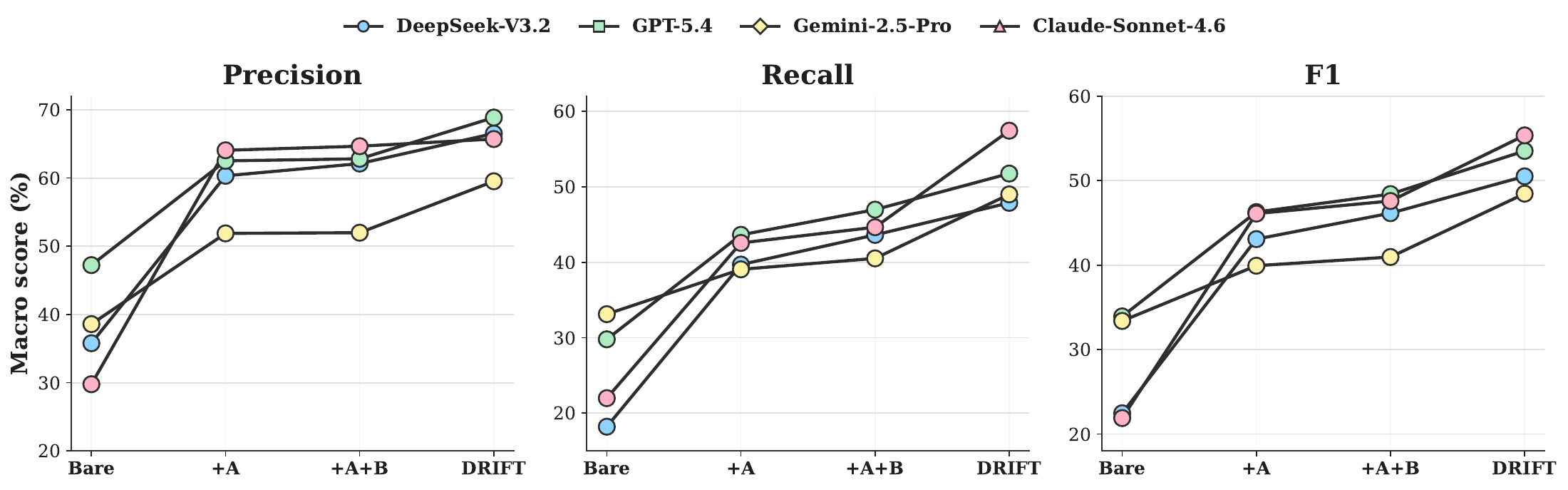}
  \vspace{-15pt}
  \caption{Ablation of Modules. Each module brings better performance.}
  \label{app:ablation_of_module}
   \vspace{-10pt}
\end{figure*}

Figure~\ref{app:ablation_of_module} reports the full module ablation across
four base models and three macro-averaged metrics. The same trend holds beyond
the main-text F1 comparison: adding the Claim Keeper yields a clear improvement
over bare prediction with the full trajectory, Support Seeker further strengthens recall by
surfacing weakly supported commitments, and the full \textsc{DRIFT} pipeline
achieves the strongest overall balance after dependency tracing. The consistency
across precision, recall, and F1 suggests that the gains are not merely caused
by over-predicting more spans, but by progressively adding structure to the
auditing process.

\clearpage
\section{Case Study}
\label{sec:case-study}

This section provides a qualitative understanding of how errors emerge and propagate in DeepResearch-style trajectories. Rather than only checking whether the final answer is correct, we inspect how each trajectory constructs intermediate commitments, how these commitments are supported or unsupported by retrieval evidence, and how early mistakes influence later reasoning. We use a unified trajectory-slice format: normal spans describe relevant but non-erroneous steps, while highlighted error spans mark the exact points where the trajectory introduces or propagates an incorrect or insufficiently supported commitment.

\begin{trajcase}{Case Study 1: Wrong-Candidate Propagation in a Snooker Retrieval Trajectory}

{\small

\textbf{Task.}
In a particular match of a snooker championship played before 2022, the winning player made one 50s break, two sixties breaks, two 70s breaks, and more than one 100s break. The losing player made more than two but less than five fifty-plus breaks, including one 50s break, more than one 70s break, and one 90s break. The losing player turned professional in a year between 2010 and 2017 exclusive, and the referee became an international referee in December of a year before 2022. The task is to identify the combined points of the players in all frames of this match.

\textbf{Gold answer.}
1419

\medskip
\textbf{Trajectory result.}
The trajectory follows a nonmatching 2021 UK Championship Final path and fails to identify the correct match. The key failure is not merely an incorrect total, but the selection of a match whose winner/loser roles and break-pattern constraints do not satisfy the prompt.

\medskip
\textbf{Annotated error summary.}

\medskip
\begingroup
\footnotesize
\setlength{\tabcolsep}{3pt}
\renewcommand{\arraystretch}{1.3}
\begin{tabularx}{\linewidth}{
@{}
>{\ttfamily}p{0.09\linewidth}
>{\raggedright\arraybackslash}p{0.27\linewidth}
>{\raggedright\arraybackslash}X
@{}
}
\textbf{\textrm{Span}} &
\textbf{Error type} &
\textbf{Annotation rationale} \\
\hline
s001 &
\makecell[tl]{\texttt{wrong\_candidate}\\\texttt{commitment}} &
The trajectory introduces the 2021 UK Championship Final as a likely candidate before validating the full break-pattern, loser-identity, professional-year, and referee constraints. \\[3pt]
s003 &
\makecell[tl]{\texttt{candidate}\\\texttt{reinforcement}} &
The trajectory strengthens the wrong candidate by calling Zhao Xintong vs. Luca Brecel a strong match and then investigates Zhao's professional year, even though the prompt requires the losing player's year. \\[3pt]
s007 &
\makecell[tl]{\texttt{entity\_role}\\\texttt{mismatch}} &
The trajectory uses Zhao Xintong's 2016 professional year to satisfy the losing-player condition, despite Zhao being the winner of the selected match. \\[3pt]
s008 &
\makecell[tl]{\texttt{error}\\\texttt{propagation}} &
The trajectory finalizes along the same invalid candidate branch even though the break pattern, loser constraint, and referee timing remain unresolved. \\
\end{tabularx}
\endgroup

\medskip
\textbf{Reasoning chain.}
\begin{enumerate}[leftmargin=1.5em, itemsep=0.25em]
\item The trajectory first explores several snooker events and players, but introduces the 2021 UK Championship Final before validating the complete conjunction of constraints.
\item It then treats Zhao Xintong vs. Luca Brecel as a strong candidate and begins extracting facts about that match.
\item The decisive local contradiction appears when Zhao Xintong's professional year is used for the losing-player constraint, although Zhao was the winner.
\item Later retrieval focuses on the referee and tries to patch the same candidate branch instead of reopening the match search.
\item The final answer fails because the trajectory never recovers from the initial wrong-candidate commitment.
\end{enumerate}

\medskip
\textbf{Local trajectory slice.}

\begin{errorspan}{s001 -- Premature candidate introduction}{wrong candidate commitment}
\footnotesize
\textbf{Trace excerpt.}
The trajectory searches professional years and break statistics, tries Kyren Wilson, Luca Brecel, and 2020 World Championship routes, then introduces the 2021 UK Championship Final as a likely candidate.

\textbf{Annotation note.}
This is the first marked error. The trajectory commits to a candidate before checking whether the candidate satisfies the complete conjunction of constraints. This early commitment narrows the subsequent search space and makes later verification steps revolve around the wrong match.
\end{errorspan}

\begin{normalspan}{s002 -- Candidate fact gathering}
\footnotesize
\textbf{Trace excerpt.}
The trajectory checks frame-by-frame scores, CueTracker statistics, and basic match details for the 2021 UK Championship Final.

\textbf{Role in chain.}
This span is not itself marked as erroneous because gathering details about a candidate can be useful. However, because the candidate was already weakly grounded, this fact-gathering step does not repair the upstream selection error.
\end{normalspan}

\begin{errorspan}{s003 -- Wrong candidate strengthened}{candidate reinforcement}
\footnotesize
\textbf{Trace excerpt.}
The trajectory explicitly calls Zhao Xintong vs. Luca Brecel a ``strong candidate'' and pursues Zhao Xintong's 2016 professional year.

\textbf{Annotation note.}
This span reinforces the incorrect branch. The professional-year check is aimed at Zhao Xintong, but the task asks for the losing player's professional year. The trajectory therefore strengthens a candidate using a fact attached to the wrong role.
\end{errorspan}

\begin{normalspan}{s004--s006 -- Referee verification branch}
\footnotesize
\textbf{Trace excerpt.}
The trajectory searches whether referee Ben Williams became an international referee in December, visiting WST profile pages, snooker-blog-style sources, and WPBSA-like evidence.

\textbf{Role in chain.}
This branch attempts to verify another constraint, but it remains anchored to the same candidate. It illustrates how a wrong candidate can redirect later retrieval toward patching a flawed hypothesis rather than testing alternatives.
\end{normalspan}

\begin{errorspan}{s007 -- Winner fact used for loser constraint}{entity-role mismatch}
\footnotesize
\textbf{Trace excerpt.}
The trajectory admits that no specific December date for Ben Williams was found, but still claims Zhao Xintong's 2016 professional year satisfies the losing-player criterion.

\textbf{Annotation note.}
This is the most visible local contradiction. Zhao Xintong was the winner in the selected match, so his professional year cannot satisfy a condition about the losing player. Bare-style systems often detect this span but miss that the mistake originates earlier.
\end{errorspan}

\begin{errorspan}{s008 -- Final continuation along invalid branch}{error propagation}
\footnotesize
\textbf{Trace excerpt.}
The trajectory finalizes along the 2021 UK Championship Final path even though the break pattern, losing-player identity, and referee constraints remain unmet.

\textbf{Annotation note.}
This span propagates the earlier candidate error into the final result. The error is not an isolated final answer mistake; it is the endpoint of a chain from premature candidate selection to role mismatch and unresolved constraint checking.
\end{errorspan}

\medskip
\textbf{Model behavior.}
Most bare-style frameworks mainly predict \texttt{s007}, detecting the explicit winner/loser mismatch but not the upstream candidate-selection failure. Frameworks with stronger trajectory awareness recover more of the propagation chain. Claude + DRIFT and DeepSeek + DRIFT exactly predict \texttt{s001,s003,s007,s008}, matching the gold annotation and treating the case as a multi-span error chain rather than a single local contradiction.

} 
\end{trajcase}

\bigskip

\begin{trajcase}{Case Study 2: Correct Final Answer with an Unsupported Evidence Chain}

{\small

\textbf{Task.}
An essay was written by a candidate for a PhD in history in 2008 on the subject of a 19th-century conflict. The acknowledgments thanked an academic who completed their undergraduate and doctoral studies on different continents. The author eventually completed their PhD at the same university at which they had completed their undergrad and went on to give seven academic invited talks and presentations on the Siege of Leningrad in 2018 and 2019 combined. The task is to identify the title of the essay.

\textbf{Gold answer.}
\emph{Heroes, Cowards, \& Traitors: The Crimean War \& its Challenge to Russian Autocracy}

\medskip
\textbf{Trajectory result.}
The trajectory gives the correct final title, but the intermediate evidence chain is not fully supported by visible retrieval evidence. We therefore count this as a trajectory-level error rather than an answer-level error: the final string is correct, but the path to it overstates what has been verified.

\medskip
\textbf{Annotated error summary.}

\medskip
\begingroup
\footnotesize
\setlength{\tabcolsep}{3pt}
\renewcommand{\arraystretch}{1.3}
\begin{tabularx}{\linewidth}{
@{}
>{\ttfamily}p{0.09\linewidth}
>{\raggedright\arraybackslash}p{0.27\linewidth}
>{\raggedright\arraybackslash}X
@{}
}
\textbf{\textrm{Span}} &
\textbf{Error type} &
\textbf{Annotation rationale} \\
\hline
s003 &
\makecell[tl]{\texttt{unsupported}\\\texttt{worker\_claim}} &
The worker reports Alexis Peri as a seven-talk match, but the visible trajectory contains no corresponding worker tool calls and several talks are not clearly specific to the Siege of Leningrad. \\[3pt]
s004 &
\makecell[tl]{\texttt{unsupported}\\\texttt{adoption}} &
The main agent adopts the worker's seven-talk claim and adds verification-style statements as if all constraints had been independently confirmed. \\[3pt]
s005 &
\makecell[tl]{\texttt{unsupported}\\\texttt{finalization}} &
The final report repeats the same seven-talk and verification claims, propagating the unsupported evidence chain into the final answer. \\
\end{tabularx}
\endgroup

\medskip
\textbf{Reasoning chain.}
\begin{enumerate}[leftmargin=1.5em, itemsep=0.25em]
\item The main agent starts from the distinctive 2018--2019 Siege of Leningrad talks condition and delegates a worker search.
\item The worker claims to have identified Alexis Peri as a seven-talk match.
\item The visible trajectory does not provide enough retrieval evidence for the full seven-talk claim.
\item The main agent nevertheless adopts the worker result and extends it with additional verification-style statements.
\item The final answer string is correct, but the trajectory overstates what has been verified.
\end{enumerate}

\medskip
\textbf{Local trajectory slice.}

\begin{normalspan}{s001 -- Constraint decomposition}
\footnotesize
\textbf{Trace excerpt.}
The main agent decomposes the question and decides to start from the distinctive 2018--2019 Siege of Leningrad talks condition, then delegates a worker search.

\textbf{Role in chain.}
This is a reasonable search strategy. The talk-count condition is distinctive and can help identify the relevant historian before checking the essay and acknowledgments.
\end{normalspan}

\begin{normalspan}{s002 -- Worker search instruction}
\footnotesize
\textbf{Trace excerpt.}
The worker subtask asks for historians, CVs, faculty pages, and event announcements that could identify a candidate with around seven relevant talks across the two years.

\textbf{Role in chain.}
This span defines a plausible retrieval plan. It is not erroneous by itself; the problem begins when the worker's later report presents the result as fully verified without visible support.
\end{normalspan}

\begin{errorspan}{s003 -- Unsupported worker report}{unsupported worker claim}
\footnotesize
\textbf{Trace excerpt.}
The worker reports a comprehensive search and identifies Alexis Peri as a seven-talk match, but the visible trajectory contains no corresponding worker tool calls and several talk titles are not clearly specific to the Siege of Leningrad.

\textbf{Annotation note.}
This is the first marked error. The issue is not that Alexis Peri is necessarily the wrong candidate, but that the worker presents the seven-talk condition as verified without enough visible evidence for that verification.
\end{errorspan}

\begin{errorspan}{s004 -- Main-agent adoption of unsupported evidence}{unsupported adoption}
\footnotesize
\textbf{Trace excerpt.}
The main agent adopts the worker's seven-talk claim and adds verification-style claims about Peri's education, the essay, the acknowledgments, and Victoria Frede's education as if they had all been independently confirmed.

\textbf{Annotation note.}
This span propagates the unsupported worker claim into the main reasoning path. The agent's wording turns a weakly supported candidate into an apparently verified chain of constraints.
\end{errorspan}

\begin{errorspan}{s005 -- Unsupported final verification claim}{unsupported finalization}
\footnotesize
\textbf{Trace excerpt.}
The final report gives the correct essay title but repeats that the seven talks and all constraints were independently verified.

\textbf{Annotation note.}
This is an evidence-chain error. The final answer string is correct, but the trajectory still contains an error because it overclaims the evidential basis for the answer.
\end{errorspan}

\medskip
\textbf{Model behavior.}
Many bare-style settings predict an empty error set, suggesting that they are strongly influenced by final answer correctness. In contrast, GPT-5.4 variants, Claude + DRIFT, and DeepSeek + DRIFT identify \texttt{s003,s004,s005}. This case shows why trajectory evaluation must inspect whether intermediate claims are supported, not only whether the final answer matches the target.

} 
\end{trajcase}

\bigskip

\begin{trajcase}{Case Study 3: Candidate-Scope Error in a Visual-Retrieval Trajectory}

{\small

\textbf{Task.}
Which of the fruits shown in Janet Fish's 2008 painting
\emph{Embroidery from Uzbekistan}
were served as part of the October 1949 breakfast menu for the ocean liner
later used as a floating prop for the film \emph{The Last Voyage}?
The answer should be a comma-separated list, ordered clockwise from the
12 o'clock position in the painting.

\textbf{Gold answer.}
\emph{pears, bananas}

\medskip
\textbf{Trajectory result.}
The trajectory incorrectly concludes that the painting does not contain bananas and treats the requested pear/banana arrangement as impossible. This conflicts with the gold annotation, which treats pears and bananas as the relevant painting fruits.

\medskip
\textbf{Annotated error summary.}

\medskip
\begingroup
\footnotesize
\setlength{\tabcolsep}{3pt}
\renewcommand{\arraystretch}{1.3}
\begin{tabularx}{\linewidth}{
@{}
>{\ttfamily}p{0.09\linewidth}
>{\raggedright\arraybackslash}p{0.27\linewidth}
>{\raggedright\arraybackslash}X
@{}
}
\textbf{\textrm{Span}} &
\textbf{Error type} &
\textbf{Annotation rationale} \\
\hline
s004 &
\makecell[tl]{\texttt{candidate}\\\texttt{scope\_error}} &
The trajectory finalizes the painting's fruit list from a non-exhaustive transcript, omitting bananas from the candidate set. \\[3pt]
s007 &
\makecell[tl]{\texttt{unsupported}\\\texttt{commitment}} &
The trajectory commits to the false conclusion that the painting does not contain bananas and declares the requested pear/banana arrangement impossible. \\
\end{tabularx}
\endgroup

\medskip
\textbf{Reasoning chain.}
\begin{enumerate}[leftmargin=1.5em, itemsep=0.25em]
\item The trajectory first identifies the ocean liner in \emph{The Last Voyage} and links it to the October 1949 breakfast menu.
\item It then tries to identify fruits in \emph{Embroidery from Uzbekistan}.
\item The first marked error occurs when the agent accepts an incomplete fruit list: watermelon, pears, and lemons.
\item Later spans continue searching for image evidence and specifically mention pears and bananas, so the conflict is visible in the local trajectory.
\item The second marked error occurs when the agent resolves that conflict incorrectly: it asserts that there are no bananas and treats the prompt as impossible.
\end{enumerate}

\medskip
\textbf{Local trajectory slice.}
Only the spans most relevant to the error chain are shown below; therefore, the numbering follows the original trajectory and is not necessarily consecutive.

\begin{normalspan}{s001 -- Task setup}
\footnotesize
\textbf{Trace excerpt.}
The agent restates the question and plans to identify the fruits in the painting, the ocean liner used in \emph{The Last Voyage}, and the October 1949 breakfast menu.

\textbf{Role in chain.}
This span preserves the original constraints: painting fruits, breakfast-menu items, and clockwise ordering.
\end{normalspan}

\begin{normalspan}{s003 -- Ocean-liner branch}
\footnotesize
\textbf{Trace excerpt.}
The trajectory searches for the ship used as a floating prop in \emph{The Last Voyage} and identifies the ocean liner as \emph{SS Ile de France}. It then moves toward the ship's history and menu evidence.

\textbf{Role in chain.}
This is a relevant retrieval branch. It establishes the menu side of the question, but it does not yet decide which painting fruits are present.
\end{normalspan}

\begin{errorspan}{s004 -- Painting fruit list}{candidate scope error}
\footnotesize
\textbf{Trace excerpt.}
The agent searches: ``What fruits are depicted in the painting Embroidery from Uzbekistan by Janet Fish?'' and uses a transcript-like source. It concludes that the fruits are ``watermelon, pears, and lemons.''

\textbf{Annotation note.}
This is the first marked error. The span treats a partial description as if it were an exhaustive fruit list. Because bananas are omitted here, the candidate set becomes too narrow before the menu-matching step.
\end{errorspan}

\begin{normalspan}{s005 -- Conflicting image evidence appears}
\footnotesize
\textbf{Trace excerpt.}
The trajectory searches the painting title again and reaches collection/archive-style pages for \emph{Embroidery from Uzbekistan}. It also searches for ``Janet Fish Embroidery from Uzbekistan 2008 image pears bananas arrangement.''

\textbf{Role in chain.}
This span shows that the banana hypothesis has not disappeared. The local evidence now conflicts with the earlier watermelon/pear/lemon-only list.
\end{normalspan}

\begin{normalspan}{s006 -- Targeted arrangement query}
\footnotesize
\textbf{Trace excerpt.}
The trajectory asks: ``What is the arrangement of pears and bananas in the painting, starting from the 12 o'clock position and going clockwise?'' It also searches image and art-site sources for the painting.

\textbf{Role in chain.}
This span explicitly frames pears and bananas as potentially relevant painting fruits. It creates a recovery opportunity: the agent could have reopened the fruit-identification step instead of treating the earlier list as exhaustive.
\end{normalspan}

\begin{errorspan}{s007 -- False impossibility claim}{unsupported commitment}
\footnotesize
\textbf{Trace excerpt.}
The agent searches for Janet Fish still-life paintings and detailed descriptions of \emph{Embroidery from Uzbekistan}. It then states that the painting does not contain bananas and that the prompt likely confused it with another painting.

\textbf{Annotation note.}
This is the second marked error. Instead of resolving the conflict by verifying the image/content evidence, the trajectory commits to a no-banana conclusion and declares the requested pear/banana answer impossible. This directly conflicts with the gold answer, \emph{pears, bananas}.
\end{errorspan}

\begin{normalspan}{s008 -- Control-flow recovery}
\footnotesize
\textbf{Trace excerpt.}
The agent notes that it needs to break the problem into smaller steps and call the search agent one at a time.

\textbf{Role in chain.}
This is a process-control recovery attempt, not itself a marked content error. The main content errors are the incomplete fruit set in \texttt{s004} and the unsupported no-bananas commitment in \texttt{s007}.
\end{normalspan}

} 
\end{trajcase}

\bigskip

\paragraph{Takeaway.}
Together, these cases show that DeepResearch errors are best understood as trajectory-level phenomena. Some failures begin with an early wrong candidate and propagate through later checks; others preserve the correct final answer but rely on unsupported intermediate evidence; still others arise from overly narrow candidate scopes inside an otherwise relevant retrieval branch. The colored trajectory-slice format makes these distinctions explicit by separating normal retrieval steps from the exact spans where incorrect or unsupported commitments are introduced and propagated.

\clearpage
\onecolumn 
\section{Prompt}

This section lists the prompts used by DRIFT and the bare evaluation baseline. Each call uses a system prompt to enforce JSON-only output and a user prompt to specify the role, task, and output schema. We omit the concrete trajectory payload for brevity and show only its placeholder fields.

\paragraph{Common system prompt.}
All modules use the same system prompt. It constrains the model to behave as a careful trajectory reader and return only a valid JSON object, which makes the outputs easier to parse and compare across methods.

\begin{promptbox}[label={prompt:system}]{Prompt 0. Common System Prompt}
[System Prompt]

You are a careful trajectory-reading assistant.
Output only one valid JSON object.
No markdown.
No extra text.
\end{promptbox}

\paragraph{Bare evaluation prompt.}
The bare baseline directly reads the full trajectory once and predicts error spans without claim decomposition, support checking, or dependency backtracing.

\begin{promptbox}[label={prompt:bare-eval}]{Prompt 1. Bare Evaluation}
DRIFT Audit Room ablation - bare model.

You are evaluating one deep-research trajectory for span-level error localization. This is a bare single-call evaluation: read the full question and all ordered spans once, then predict final error spans directly.

Mark a span only if the span itself contains a committed harmful mistake, an unsupported committed conclusion, a harmful premature finalization, or a harmful continuation. Do not mark harmless exploration, ordinary evidence gaps, isolated tool failures without commitment, retries, search queries, tentative candidate pivots, or generic uncertainty.

Prefer a sparse set of committed harmful spans. If the actual harmful commitment appears only in the final report, output only that final span. If an early span already commits to the wrong answer path or harmful no-answer decision, mark that earliest committed span and any later spans that explicitly rely on or finalize it. If there is no committed harmful error, return an empty error_span_ids list.

Return JSON only.

Schema:
{
  "traj_id": "...",
  "error_span_ids": ["s004", "s007"],
  "earliest_harmful_span_id": "s004",
  "reasons": [
    {
      "span_id": "s004",
      "reason": "short string"
    }
  ]
}

Input:
{
  "question": "...",
  "traj_id": "...",
  "spans": [
    {
      "span_id": "s001",
      "span_text": "..."
    }
  ]
}
\end{promptbox}

\paragraph{A: Claim Keeper.}
Claim Keeper converts the trajectory into an audit ledger. It identifies consequential claims and records when they become commitments used by later reasoning, but it does not decide final error spans.

\begin{promptbox}[label={prompt:claim-keeper}]{Prompt 2. A: Claim Keeper}
DRIFT Audit Room - A: Claim Keeper.

Read the question and ordered trajectory as an audit ledger, not as a final judge. Track consequential claims the agent comes to believe: entities, constraints, dates/ranges, evidence interpretations, retrieval coverage, computations, and process/tool assumptions.

For each claim, record when it first appears, when it first becomes consequential for later work, and which later spans use it. Keep only decision-critical claims: claims that choose the answer path, narrow to one candidate, verify a hard constraint, justify the final response, or explain why no answer can be produced.

A search query, subtask request, tool call, or candidate name inside a query is not a commitment. Record it only if the span also says it has identified/found/verified the candidate or uses it as the answer path. Use status=finalized only for claims that are submitted, finalized, used in the final answer/no-answer, or explicitly treated as solved. Use status=consequential for claims that drive later work but are not yet final. Use tentative/exploratory for probes and candidates.

For no-answer cases, track the claim that the agent cannot answer only when it says final answer / cannot determine / unable / apology and stops or avoids the requested computation. Keep the ledger compact: prefer 3-5 claims, and never include more than the few claims that could change the final error decision.

Do not decide final error spans. Return JSON only.

Schema:
{
  "traj_id": "...",
  "task_goal": "short string",
  "hard_constraints": ["short string"],
  "claims": [
    {
      "claim_id": "c1",
      "claim": "short factual or procedural claim",
      "introduced_at": "s003",
      "becomes_consequential_at": "s011",
      "used_by": ["s011", "s015"],
      "claim_type": "entity",
      "status": "tentative"
    }
  ],
  "notes": "short string"
}

Input:
{
  "question": "...",
  "traj_id": "...",
  "spans": [
    {
      "span_id": "s001",
      "span_text": "..."
    }
  ],
  "allowed_claim_types": ["entity", "constraint", "evidence", "retrieval", "compute", "process"],
  "allowed_status_values": ["exploratory", "tentative", "consequential", "finalized"]
}
\end{promptbox}

\paragraph{B: Broad Support Seeker.}
Support Seeker checks whether the claims found by Claim Keeper are actually supported by the trajectory. It builds high-recall claim-support links and flags weak, missing, or conflicting support, but still does not output final error spans.

\begin{promptbox}[label={prompt:support-seeker}]{Prompt 3. B: Broad Support Seeker}
DRIFT Audit Room - B: Broad Support Seeker.

Your role is high-recall candidate/support mining, not final judgment. For each consequential claim in A's ledger, collect the spans that appear to support it and expose every decision-critical support risk that may need specialist checking.

Use direct only when the trajectory explicitly verifies the claim's decisive identity, constraint, source link, count/computation, retrieval coverage, or no-answer justification. Use weak when there is related evidence but one decisive link is only implied, assumed, based on a snippet, based on a partial comparison, or not checked against the exact question.

Use missing when the trajectory commits to a final answer/no-answer/computation but shows no support for a required decisive link. Use conflicting when shown evidence contradicts the claim, the agent ignores a hard constraint, or a tool/source result undermines the commitment.

Route weak/missing/conflicting claims to the most relevant auditors. This broad B step is allowed to over-route because C will later verify/filter the candidates. Preserve A's claim semantics exactly. Do not invent new claims, rewrite answer claims into generic process claims, or mark final errors yourself.

Return JSON only.

Schema:
{
  "traj_id": "...",
  "support_records": [
    {
      "claim_id": "c1",
      "support_spans": ["s003", "s008"],
      "support_status": "weak",
      "missing_support": "short string",
      "needs_auditors": ["entity", "evidence"]
    }
  ],
  "notes": "short string"
}

Input:
{
  "question": "...",
  "traj_id": "...",
  "claim_ledger": {},
  "spans": [
    {
      "span_id": "s001",
      "span_text": "..."
    }
  ],
  "allowed_support_status": ["direct", "weak", "missing", "conflicting"],
  "allowed_auditors": ["entity", "constraint", "evidence", "retrieval", "compute", "process"]
}
\end{promptbox}

\paragraph{C: Specialist Auditor Gate.}
Specialist Auditors are routed to narrow claim-support questions. Each auditor checks one type of possible failure, such as entity matching, constraint satisfaction, evidence use, retrieval coverage, computation, or process/tool reliability, and returns a typed chain edit rather than final error labels.

\begin{promptbox}[label={prompt:specialist-auditor}]{Prompt 4. C: Specialist Auditor Gate}
DRIFT Audit Room - C: Specialist Auditor Gate ({auditor}).

B deliberately over-routes weak support risks. Your job is to produce a typed patch over B's exposed claim chain, not to find unrelated errors. Think like a margin editor on A+B's draft: KEEP the candidate if it is a real harmful commitment, DROP it if it is clearly diagnostic context rather than an error, or ADD/relocate to the true commitment span only within the exposed neighborhood.

Return supported when the relevant spans directly establish the decisive link for this claim. Return insufficient_but_nonharmful when B found a real weakness but the claim is exploratory, abandoned, later corrected, not used by the final answer/no-answer, or only suffers from incomplete citation/logging. Return harmful_unsupported_commitment when the claim is consequential/finalized, the decisive link remains weak or missing, and the span commits to an answer path, final answer, computation, or no-answer decision. Return conflicting_support when shown evidence contradicts the claim or a hard constraint is violated.

Use support_assessment to record whether support is direct, weak, missing, or conflicting. Use chain_action=confirm to keep A+B's span/claim, remove_false_positive to drop a clearly non-error span, confirm_or_add to add or relocate only to an exposed commitment span, and no_change when evidence is ambiguous. Do not choose pure searches, tool calls, snippets, failed retrieval attempts, broad plans, or candidate mentions as responsible_span unless the span itself states the claim as settled.

Return JSON only.

Schema:
{
  "traj_id": "...",
  "claim_id": "c1",
  "auditor": "evidence",
  "support_assessment": "weak",
  "commitment_status": "committed",
  "impact_status": "finalized",
  "decisive_defect": "required_evidence_failed",
  "failure_mechanism": "unsupported_inference",
  "verdict": "harmful_unsupported_commitment",
  "responsible_span": "s003",
  "confidence": "high",
  "chain_action": "confirm_or_add",
  "why": "short string",
  "follow-up_span_ids": ["s011", "s015"]
}

Input:
{
  "question": "...",
  "traj_id": "...",
  "auditor": "entity",
  "claim": {},
  "support_record": {},
  "relevant_spans": [
    {
      "span_id": "s003",
      "span_text": "..."
    }
  ],
  "allowed_verdicts": ["supported", "harmful_unsupported_commitment", "conflicting_support", "insufficient_but_nonharmful"],
  "allowed_support_assessment": ["direct", "weak", "missing", "conflicting"],
  "allowed_commitment_status": ["none", "exploratory", "tentative", "committed", "finalized"],
  "allowed_impact_status": ["none", "local_only", "follow-up_used", "finalized", "harmful_blocking"],
  "allowed_decisive_defects": ["none", "explicit_contradiction", "hard_constraint_violation", "wrong_entity_or_mapping", "fabricated_or_misused_evidence", "required_evidence_failed", "premature_or_incomplete_finalization", "blocking_no_answer", "unsupported_but_no_decisive_defect"],
  "allowed_failure_mechanisms": ["none", "wrong_mapping", "constraint_violation", "evidence_misuse", "unsupported_inference", "premature_commitment", "retrieval_drift", "compute_error", "tool_failure_used_as_basis"],
  "allowed_chain_actions": ["confirm", "remove_false_positive", "confirm_or_add", "no_change"],
  "allowed_confidence": ["low", "medium", "high"]
}
\end{promptbox}

\paragraph{Final dependency backtrace.}
The final Dependency Tracer step closes the audit loop. It starts from the broad A+B candidate chain and uses specialist gate verdicts to distinguish committed error spans from suspicious but non-error spans.

\begin{promptbox}[label={prompt:dependency-tracer}]{Prompt 5. A: Dependency Backtrace with C Gate}
DRIFT Audit Room - A: Dependency Tracer with C Gate v3.

Broad B has already produced a high-recall candidate chain. C should correct B mainly by filtering false alarms. Therefore this final reducer is conservative about adding new spans and careful about preserving the prior first/follow-up chain.

Start from prior_support_trace.error_span_ids as the working set. Preserve prior_support_trace.first_error_span unless the span is clearly only a search query, tool call, snippet, retry, broad plan, or abandoned candidate with no commitment language.

Preserve prior follow-up spans that are also listed in high-confidence C follow-up_span_ids, because C has confirmed they continue the same unsupported claim. Remove a prior span only when C clearly classifies the same claim as supported/insufficient_but_nonharmful and no high-confidence unsupported/conflicting verdict for that claim lists or explains the span. Remove prior spans that are pure retrieval/process noise, but do not remove repeated settled-claim spans just because the chain could be shorter.

Do not add new spans by default. A new span may be added only if C has high confidence and the span is a final answer/no-answer, an explicit first commitment, an explicit false verification, or a completed computation/count/source claim. If C finds an earlier error but Broad B already has a correct later committed span, add the error only when it improves earliest-error localization and the earlier span itself commits to the same harmful claim.

For no-answer cases, keep the explicit cannot-answer/stop/final no-answer commitments and any prior follow-up spans that C confirms; do not add every failed search. For final answer cases, keep the prior committed chain unless C proves parts are non-errors. If only the final report commits, output only the final report span. If uncertain whether a change improves the prior trace, keep the prior trace unchanged.

Return JSON only.

Schema:
{
  "traj_id": "...",
  "first_error_span": "s003",
  "follow-up_error_spans": ["s011", "s015"],
  "not_errors": ["s004", "s005"],
  "reason": "short string",
  "error_span_ids": ["s003", "s011", "s015"]
}

Input:
{
  "question": "...",
  "traj_id": "...",
  "claim_ledger": {},
  "support_records": [],
  "audit_verdicts": [],
  "prior_support_trace": {},
  "relevant_spans": [
    {
      "span_id": "s001",
      "span_text": "..."
    }
  ]
}
\end{promptbox}

\end{document}